\documentclass[lettersize,journal]{IEEEtran}
\usepackage{amsmath,amsfonts}
\usepackage{algorithm2e}
\RestyleAlgo{ruled}
\usepackage{array}
\usepackage[caption=false,font=footnotesize,labelfont=sf,textfont=sf]{subfig}
\usepackage{textcomp}
\usepackage{stfloats}
\usepackage{url}
\usepackage{verbatim}
\usepackage{graphicx}
\usepackage{multicol}
\usepackage{svg}
\usepackage[style=ieee, citestyle=numeric-comp, backend=biber, url=true]{biblatex}
\AtEveryBibitem{
    \clearfield{urlyear}
    \clearfield{urlmonth}
}
\usepackage[T1]{fontenc}

\usepackage[utf8]{inputenc}
\usepackage{picinpar}
\usepackage{pdfpages}
\usepackage{amsmath}
\usepackage{url}
\usepackage{flushend}
\usepackage{colortbl}
\usepackage{soul}
\usepackage{multirow}
\usepackage{pifont}
\usepackage{color}
\usepackage{alltt}
\usepackage[hidelinks]{hyperref}
\usepackage{enumerate}
\usepackage{siunitx}
\usepackage{breakurl}
\usepackage{epstopdf}
\usepackage{pbox}
\usepackage{makecell}
\usepackage{booktabs} 
\usepackage{multirow}  
\usepackage{makecell,rotating,diagbox}
\usepackage{url}  
\graphicspath{{}}
\usepackage{bm}
\usepackage{threeparttable}
\usepackage{gensymb}
\usepackage{xargs}   
\usepackage{xcolor} 
\usepackage{balance}

\usepackage{pifont}

\SetKwComment{Comment}{/* }{ */}
\usepackage{mathtools}
\usepackage{sidecap}
\usepackage{epstopdf}

\begin{document}

\title{Curriculum-Based Reinforcement Learning for Quadrupedal Jumping: A Reference-free Design}


\author{Vassil Atanassov*, Jiatao Ding*, Jens Kober, Ioannis Havoutis, Cosimo Della Santina
\thanks{{Vassil Atanassov and Ioannis Havoutis are with the Oxford Robotics Institute, Department of Engineering Science, University of Oxford, U.K (emails: \{vassilatanassov, ioannis\}@robots.ox.ac.uk).  Jiatao Ding, Jens Kober and Cosimo Della Santina are with the Department of Cognitive Robotics, Delft University of Technology, Building 34, Mekelweg 2, 2628CD, Delft, The Netherlands (e-mails: \{J.Ding-2, C.DellaSantina, j.kober\}@tudelft.nl). 
   Cosimo Della Santina is also with the Institute of Robotics and Mechatronics, German Aerospace Center (DLR), 82234 Wessling, Germany (e-mail: cosimodellasantina@gmail.com).
}
  }
  \thanks{* Vassil Atanassov and Jiatao Ding are the corresponding authors.}
}
\maketitle
\begin{abstract}
Deep reinforcement learning (DRL) has emerged as a promising solution to mastering explosive and versatile quadrupedal jumping skills. However, current DRL-based frameworks usually rely on pre-existing reference trajectories obtained by capturing animal motions or transferring experience from existing controllers. This work aims to prove that learning dynamic jumping is possible without relying on imitating a reference trajectory by leveraging a curriculum design. Starting from a vertical in-place jump, we generalize the learned policy to forward and diagonal jumps and, finally, we learn to jump across obstacles.
Conditioned on the desired landing location, orientation, and obstacle dimensions, the proposed approach yields a wide range of omnidirectional jumping motions in real-world experiments. Particularly we achieve a 90cm forward jump, exceeding all previous records for similar robots reported in the existing literature. Additionally, the robot can reliably execute continuous jumping on soft grassy grounds, which is especially remarkable as such conditions were not included in the training stage.
\end{abstract}

\textbf{Note:} A supplementary video can be found on: \href{https://www.youtube.com/watch?v=nRaMCrwU5X8}{https://www.youtube.com/watch?v=nRaMCrwU5X8}. The code associated with this work can be found on: \href{https://github.com/Vassil17/Curriculum-Quadruped-Jumping-DRL}{https://github.com/Vassil17/Curriculum-Quadruped-Jumping-DRL}.

\section{Introduction}
Through millions of years of evolution, legged animals have adapted to loco-mote in highly complex and discontinuous environments that widely exist in nature. Goats, for example, are capable of scaling nearly vertical mountainsides and jumping across chasms several times their body length.  
While many works have tackled dynamic locomotion recently \cite{hwangbo_learning_2019,miki_learning_2022,agarwal_legged_2022}, achieving such complex controlled behaviour is still an open challenge.

Quadrupedal jumping has traditionally been investigated through model-based control, where an accurate model of the dynamical system is needed to generate optimal control inputs
\cite{nguyen_optimized_2019,nguyen_contact-timing_2022,bjelonic2022offline,ding2023robust}. In addition, these methods rely on various heuristics necessary to render the approach feasible, which limit the search space and result in conservative 
performance. 

 In contrast to model-based optimisation, model-free reinforcement learning (RL) has emerged as an effective alternative that does not require expert knowledge for control engineering and tedious gain tuning. 
 Especially, deep RL (DRL) has shown impressive generalisation and robustness capabilities in executing locomotion tasks \cite{hwangbo_learning_2019, kumar_rma_2021, rudin_learning_2022,lee_learning_2020,miki_learning_2022}. 
For quadrupedal jumping, a series of correct actions need to be taken for the robot to succeed. Paired with an inherently sparse reward structure (the robot has either jumped or not), it is exceptionally hard for the robot to learn, as most of its trials will fail. Current RL approaches tackle this by directly transferring skills from demonstrations \cite{li_learning_2022,li_robust_2023} or optimal controllers \cite{bellegarda_robust_2021,fuchioka_opt-mimic_2022, yu_dynamic_2022}. However, balancing the degree to which the agent should imitate the demonstration and generalise to new tasks is challenging and remains an open question.


\begin{figure*}[!t]
    \centering
    \includegraphics[width=\textwidth]{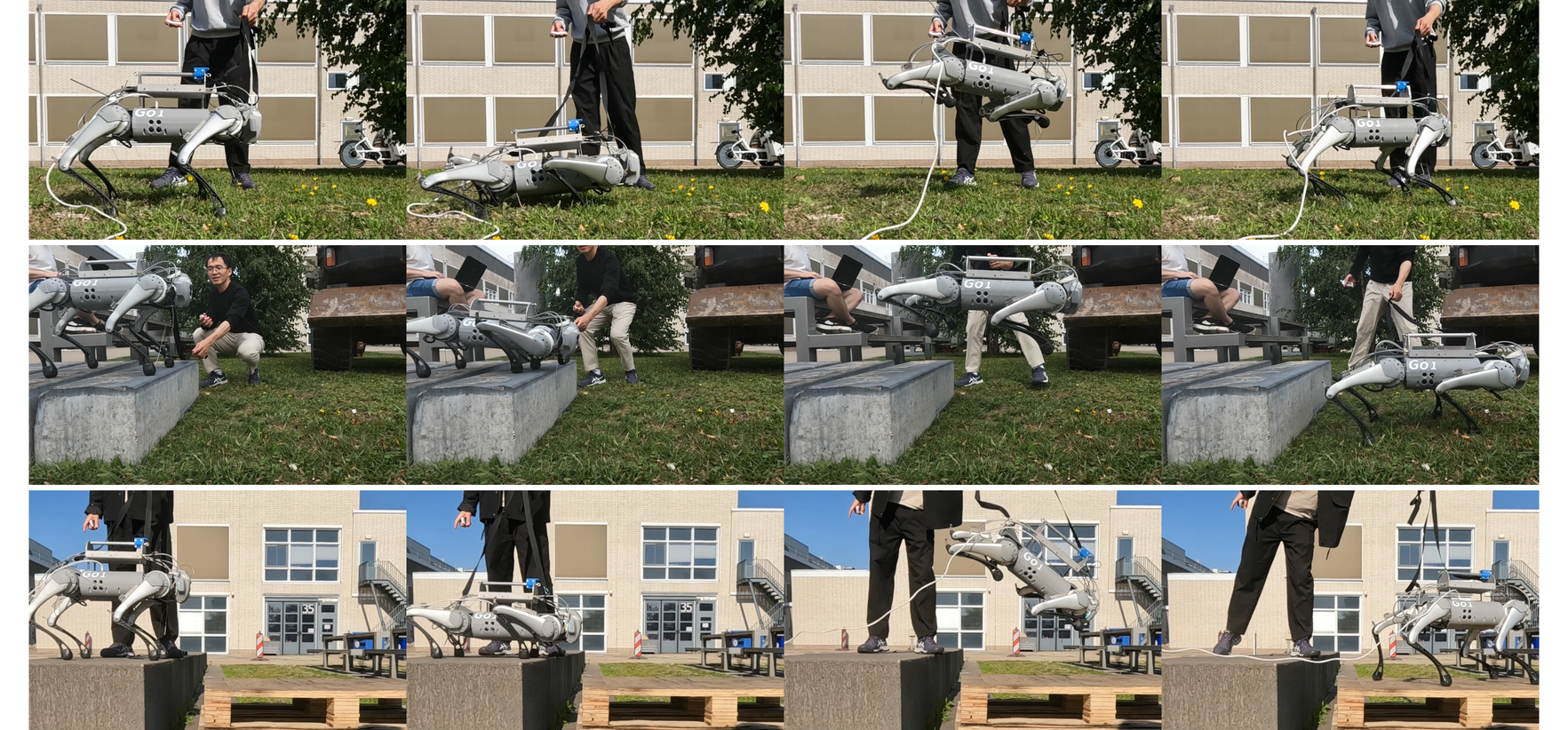}
    \vspace{-6mm}
    \caption{The Go1 robot jumps across grassland (top), jumps down onto grassland (middle) and jumps across a gap onto a lower box (bottom).}
    \label{fig:outdoors_main}
\end{figure*}

In this work, we push robots to learn to jump on their own by combining curriculum learning with DRL, eliminating the reliability of pre-computed motion references. By conditioning the policy on the desired landing location and orientation, our approach produces versatile jumping motions with just one single policy. Furthermore, by incorporating partial knowledge of the obstacles surrounding it, the robot learns different manoeuvres adapted to complex real-world scenarios.

The main contributions are summarised as follows:
\begin{itemize}
    \item We propose a curriculum-based DRL framework, which is capable of learning jumping motions without requiring motion capture data or a reference trajectory.
    \item We generalise across a wide range of jumps with a single policy for both indoor and outdoor environments. With our method, the real robot can jump 90cm forward, which, to the best of our knowledge, is the longest distance achieved on quadrupeds of a similar size. It has been demonstrated that continuous jumping across grassland and robust jumping across uneven terrains can be achieved in a zero-shot manner.
    \item We incorporate partial environmental information into the learning stage, which allows the robot to jump over more complex terrains. 

\end{itemize}

In Section~\ref{related_work}, we introduce the existing RL-based jumping controllers. In Section~\ref{curriculum_design} and Section~\ref{drl_formulation}, we separately present the curriculum design and DRL formulation. After extensively evaluating our method in Section~\ref{experiment_validation}, we discuss our approach and directions for future work in Section~\ref{conclusion_part}. 

\section{Related work}\label{related_work}
\subsection{Reinforcement learning for quadrupedal jumping}


DRL is a promising solution for accomplishing jumping tasks by offloading the computational complexity to offline training. 
One approach to learning quadrupedal jumping is by learning from demonstrations, such as from trajectories generated through optimal control \cite{bellegarda_robust_2021,fuchioka_opt-mimic_2022}, or hand-tuned reference motions \cite{li_learning_2022,li_robust_2023}. To address the challenges associated with the selection of relevant states to mimic and manage conflicting objectives, generative adversarial imitation learning (GAIL) has recently been widely adopted \cite{peng_amp_2022,escontrela_adversarial_2022,vollenweider_advanced_2022}, even when dealing with partially incomplete demonstrations \cite{li_learning_2022}. In \cite{smith_learning_2023} transfer learning is used to learn policies capable of diverse agile motions from a database of existing RL and model-based controllers. 
However, most imitation-based methods have so far shown a limited generalisation capability beyond the imitation domain. Furthermore, many of the aforementioned works rely on learning a separate policy for each unique type of motion, rather than a common task- or goal-conditioned policy.

To reduce the dependency on a motion prior, \cite{yin_discovering_2021} use a variational auto-encoder (VAE) to encapsulate motion capture data into a latent space and then combine it with a Bayesian diversity search to discover viable take-off states. 
\cite{margolis_learning_2021} trained a high-level motion planning module to produce desired centre of mass (CoM) trajectories for small hops, conditioned on visual inputs and then tracked by a model-based controller. In \cite{bellegarda_robust_2021}, deviations to reference trajectories generated by a non-linear optimal trajectory \cite{nguyen_contact-timing_2022} are learned, providing better generalisation to out-of-training domains. Similarly, \cite{yang_continuous_2023} learn action residuals to a model-based controller to achieve continuous jumping. Another work focusing on continuous hopping \cite{yang_cajun_2023} uses a learned centroidal policy to output desired centre of mass trajectories, which are tracked by a quadratic-programming-based(QP) ground reaction force (GRF) controller. Rudin et al. \cite{rudin_cat-like_2022} show cat-like jumping in low gravity by using a more complex reward function, without imitating motion clips. However, this approach has not yet been verified on Earth-like gravitational conditions. 
Recently, Vezzi et al. \cite{vezzi_two-stage_2023} proposed learning to jump by combining a first-stage evolution strategy with a second-stage DRL. Compared to \cite{vezzi_two-stage_2023}, our approach is less complex by using proximal policy optimization (PPO) \cite{schulman2017proximal} for all curriculum stages, and is capable of executing jumps conditioned on the desired jumping length and orientation.


\subsection{Curriculum learning in dynamic quadrupedal locomotion}

Curriculum learning (CL) is a training framework which progressively provides more challenging data or tasks as the policy improves. As the name suggests, the idea behind the approach borrows from human education, where complex tasks are taught by breaking them into simpler parts. 

In legged locomotion, CL has seen wide use, mainly in terms of terrain adaptation. Xie et al. \cite{xie_allsteps_2020} show how an adaptive curriculum can be used to learn stepping stone skills much more efficiently than other methods like uniform sampling. Similarly, other automatic curriculum learning methods have been proposed to vary environmental parameters based on the performance of the agents \cite{lee_learning_2020}, rather than using a manually specified curriculum. On the rewards side, Hwangbo et al. \cite{hwangbo_learning_2019} employ a curriculum which scales down certain rewards at the start. This design allows the policy to first learn how to locomote and only afterwards to be polished to satisfy the additional constraints and limits of the problem. In \cite{cheng_extreme_2023},parkour locomotion skills are learned through a well-designed terrain curriculum with a single policy, which is then distilled to a exteroception-conditioned policy. Similar parkour skills are acquired in \cite{hoeller_anymal_2023}, but the method requires separate policies for each skill, as well as a perception and navigation network, which greatly increases the computational complexity. Barkour \cite{caluwaerts_barkour_2023} uses a similar approach, but distils the specialist controllers into a single generalist transformer policy.
To learn dynamic parkour skills, \cite{zhuang_robot_2023} adopt a two-stage curriculum, transitioning from soft to hard dynamic constraints in the second stage. 
Recently, \cite{li_robust_2023} used multi-stage training to learn imitation-based vertical jumping, and then transferred that knowledge to forward jumping. While similar to our approach, however, there are a couple of significant differences - we do not require any reference trajectories, and we learn a single policy for versatile tasks.
\section{Curriculum design}\label{curriculum_design}
Defining and constraining the behaviour of jumping across specific distances is challenging as it combines two distinct behaviours: that of "jumping" and that of reaching a desired spatial point. Furthermore, an easily learnable local optimum exists, where the robot could simply walk (or crawl) toward the target point without actually jumping. To avoid converging to such undesired behaviour we use curriculum learning to decompose the problem into several simpler sub-tasks. 

\begin{figure}
    \centering
    \includegraphics[width=\columnwidth]{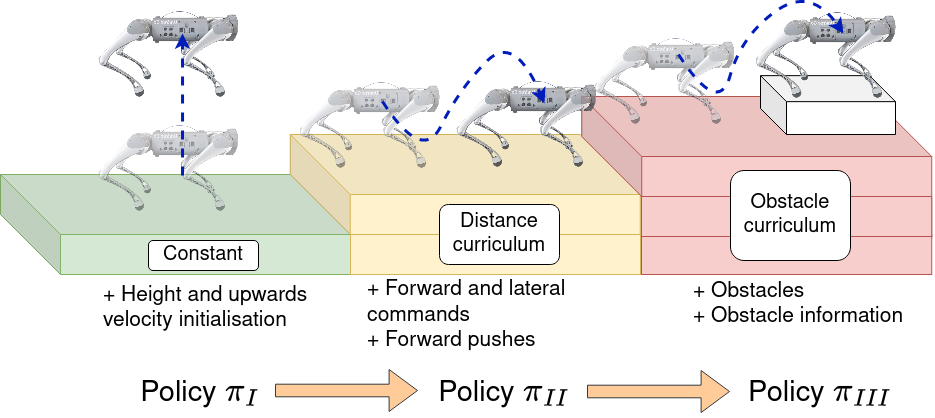}
    \vspace{-6mm}
    \caption{The curricula: jumping in place (left), long-distance jump (middle) and long-distance jump with obstacles (right). The latter two vary the jump distance/orientation and obstacle height, respectively. }
    \label{fig:curriculum}
\end{figure}

%

In our approach, we adopt two types of curriculum - on a local difficulty level and on a task level, as can be seen in Fig. \ref{fig:curriculum}. The former involves progressively (and automatically) making the environment more complex as the agent succeeds. In particular, upon successful jumps, we increase the range of desired jumping distances and obstacle heights that we sample from. The task-level curriculum is, on the other hand, manually selected and consists of training the agent for a certain number of steps at a given task. After mastering the easier jumping skill, the policy is loaded onto the next task, which might be defined differently and contains a new set of rewards. 

In the remainder , we describe each of these task-level and difficulty curricula in the progressive order of training.

\subsection{Stage I: Jumping in place} 
\label{subsec:stage_1_curriculum}
Vertical jumping without traversing a certain horizontal distance, i.e. jumping in place, is the basic component of agile jumping.
However, the lack of reference results in a learning problem with sparse rewards, given that the agent needs to first learn certain behaviours (e.g., squatting down and then pushing hard against the ground to take off) before it can reach the reward-rich states, i.e. being high in the air. As the robot does not experience these jumping-specific rewards initially, it is prone to converging to a local optimum, such as standing in place, where small rewards are collected safely.

To avoid getting stuck in this local optimum behaviour, we adopt a modified form of the reference state initialisation (RSI) technique \cite{peng_deepmimic_2018}. In imitation learning, RSI initialises the agent at random points of the reference trajectory, allowing the agent to explore such reward-rich states before it has learned the actions necessary to reach them. 
As we do not use a reference trajectory, we instead modify RSI to sample a random height and upward velocity from a predefined range. 

 \subsection{Stage II: Long-distance jump}
\label{subsec:stage_2_curriculum}
Once the robot has converged to a jumping-in-place behaviour, we further train it to perform precise forward and diagonal jumps. The first part of the command vector $\mathbf{g}\in\mathcal{R}^{13}$ (see Fig.~\ref{fig:observation_space}) in the observations specifies the desired landing point and orientation to create a goal-conditioned policy. Similarly to the \textbf{jumping in place} sub-task, 
we also adopt a curriculum-style sampling for desired landing points, where successful agents are progressed to more difficult environments where the desired jumping distance and landing yaw are sampled from a greater range.
 

\subsection{Stage III: Long-distance jump across obstacles} 
\label{subsec:stage_3_curriculum}
Finally, we introduce obstacles in the environment.  Without loss of generality, we choose three classes of obstacles, including thin barrier-like objects, box-shaped obstacles and slopes. Depending on the desired landing pose, the obstacle location and the type, the agent needs to either jump onto or over it. While it is possible to learn a general behaviour that can accomplish this without any exteroception, such behaviour will be conservative, sub-optimal and potentially much less robust. 
With this in mind, we incorporate information about the distance to the centre of the obstacles and its general dimensions (length, width, and height). In the real world, we manually specify these parameters\footnote{A separate module that estimates obstacle dimensions could be utilised. One future work would be linking exteroceptive sensors to the policy and removing the parameterisation of the world around the robot.}.

Similarly to the previous stage, we start with obstacles of smaller height. Then, successful robots progress towards more challenging terrains, whereas failing ones are demoted to easier environments. To ensure that the robot remembers the previously learned behaviour we also randomly send a certain percentage of robots to jump on flat ground, as in Stage II.

\begin{figure}
    \centering
    \includegraphics[width=\columnwidth]{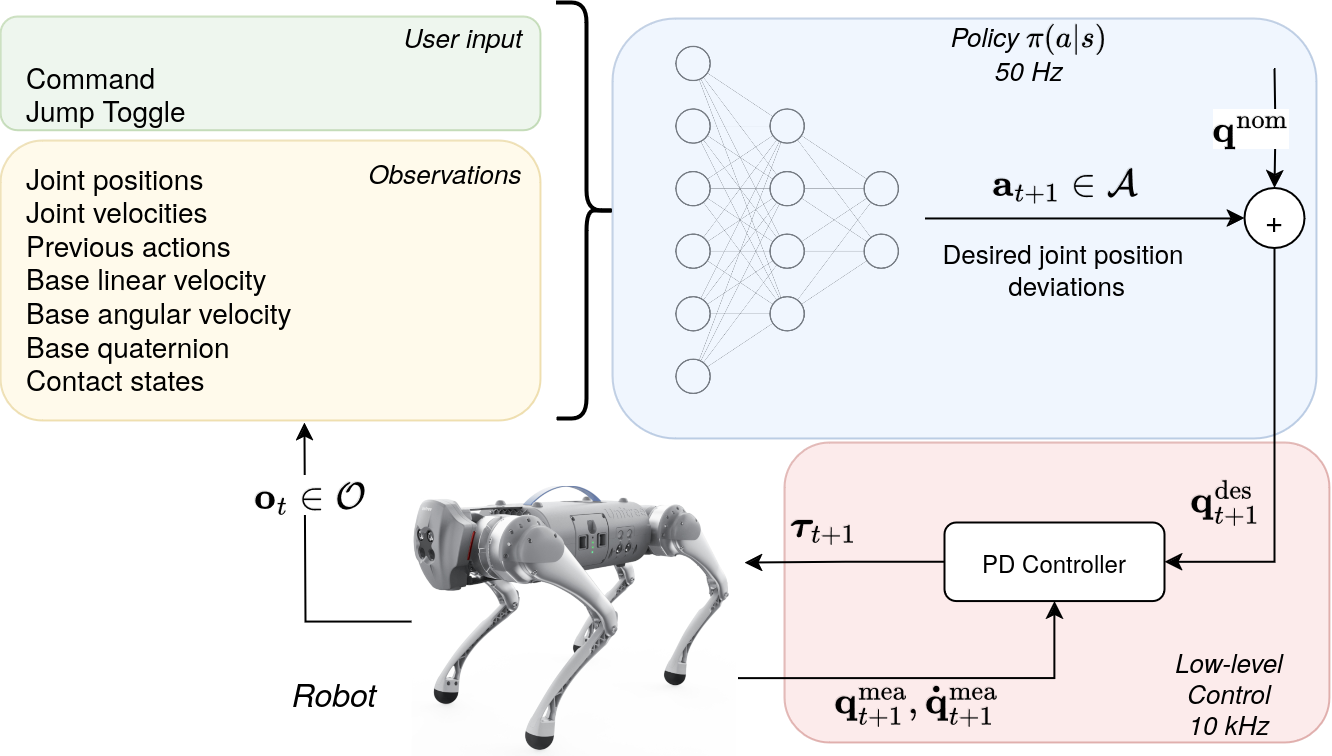}
    \vspace{-6mm}
    \caption{Control diagram of the system. The observations $\mathbf{o}_t$ include user command (in green) and a history of system states (in yellow). The policy is parameterised by a neural network (shown in blue). The output actions $\mathbf{a}_{t+1}$ are added to the nominal joint angles $\mathbf{q}^{\mathrm{nom}}$. The desired joint angles are then tracked via a PD controller which computes torque commands.}
    \label{fig:controller}
\end{figure}

\section{DRL formulation}\label{drl_formulation}

This section details the DLR formulation, as illustrated in Fig.~\ref{fig:controller}. First, preliminaries are introduced. Then, we define the key components of goal-conditioned RL, including observations, actions and reward functions. Finally, we introduce our domain randomisation scheme to mitigate the sim2real gap. 

\subsection{Preliminaries}

RL infers a policy $\pi(a_t|s_t)$ of how to act by constantly interacting with the environment. The RL problem is typically formulated as a Markov decision process (MDP), where at each step the agent interacts with the environment by taking an \textbf{action} $\textbf{a}_t \in \mathcal{A}$. Subsequently, it receives the new states of the environment $\textbf{s}_{t+1} \in \mathcal{O}$ in the form of \textbf{observation}, and the associated \textbf{reward} $\mathcal{R}_t$ that it has earned. Based on the observed state $s_{t+1}$ and its policy $\pi(a_{t+1}|s_{t+1})$ the agent can then choose a new action $a_{t+1}$. In this way, the RL algorithm optimises behaviours that yield high rewards. 
In goal-conditioned RL, the action policy can also be conditioned on specific goals, i.e. $\pi(a_t|s_t,g)$. Such a policy can be used to produce diverse behaviours depending on the specific command $g$, enabling the learning of multiple distinct behaviours under a single policy.

In this work, we formulate the following objective: finding a policy $\pi(a|s,g)$ which maximises the cumulative sum of rewards earned over the task duration. As often immediate rewards are more valuable than rewards in the distant future, a discount factor $\gamma \in (0,1]$ is commonly used. Mathematically, the full objective of maximising the sum of discounted rewards $J$, known as the return, can be written as:
\begin{equation}
    \mathop{\arg\max\limits_{\mathbf{\pi}}} \quad J(\pi) = \mathbb{E}_{\tau \sim p^\pi(\tau)}\left[\sum_{t=0}^{T}\gamma^t R_t \vert s_0 = s\right],
\end{equation}
where $R_t$ is the immediate reward at time $t$ and $s_0$ is the initial state. The expectation of the return is taken over a trajectory $\tau$ sampled by following the policy.

\subsection{Observation and action space}
\begin{figure}
    \centering
    \includegraphics[width=\columnwidth]{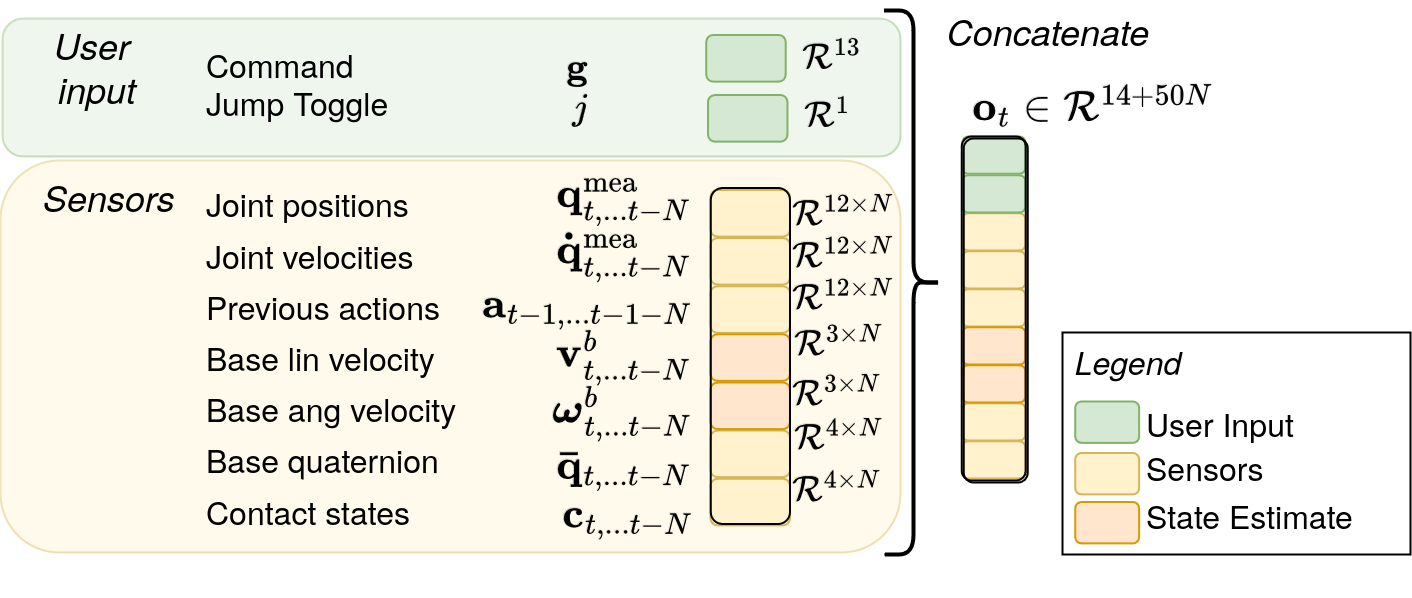}
    \vspace{-8mm}
    \caption{The definition of observations. The command $\mathbf{g}$ and jump toggle \textit{j} are provided by the user, while the remaining observations are either directly read from the sensors, or estimated using sensory data.}
    \label{fig:observation_space}
\end{figure}
\textbf{Observation space:} Using a memory of previous observations and actions allows the agent to implicitly reason about its own dynamics and the interaction with the environment \cite{hwangbo_learning_2019,lee_learning_2020}. Here, 
we use a concatenated history of the last \textit{N} steps as input to the policy
\footnote{In practice, we found that using the last 20 steps is sufficient for the task while also being fast for training.}. As illustrated in Fig. \ref{fig:observation_space}, the observation space consists of the historical base linear velocity $\mathbf{v} \in \mathbb{R}^{3\times N}$, base angular velocity $\boldsymbol{\omega} \in\mathbb{R}^{3 \times N}$ (both in the base frame), joint position $\mathbf{q} \in \mathbb{R}^{12 \times N}$, joint velocity $\mathbf{\dot{q}} \in \mathbb{R}^{12 \times N}$, previous actions $\mathbf{a}_{t-1} \in \mathbb{R}^{12 \times N}$, the base orientation (as a quaternion) $\mathbf{\bar{q}} \in \mathbb{R}^{4 \times N}$ and the foot contact states $\mathbf{c} \in \mathbb{R}^{4 \times N}$. 

\begin{figure}
    \centering
    \includegraphics[width=\columnwidth]{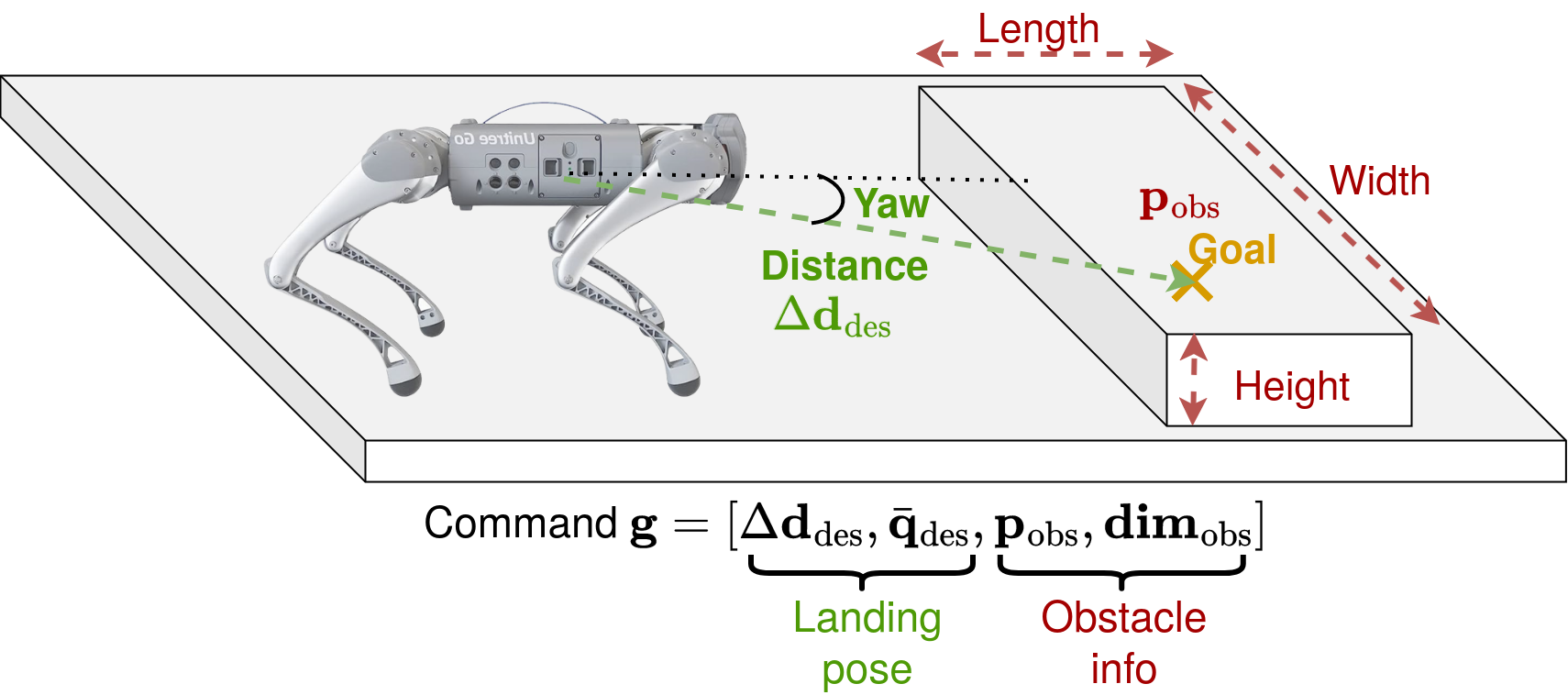}
    \vspace{-8mm}
    \caption{The command vector $\mathbf{g}$ for a forward jump onto an obstacle. In the first two training stages ($\pi_I$ and $\pi_{II}$), where no obstacles are considered, the information of the obstacle is set to zero. }
    \label{fig:forward_with_obstacles}
\end{figure}

Note that our policy is also conditioned on the command $\mathbf{g} \in \mathbb{R}^{13}$ and jump toggle $j \in \{0,1\}$, see the green block in Fig.~\ref{fig:observation_space}. As illustrated in Fig.~\ref{fig:forward_with_obstacles}, the command $\mathbf{g} \in [\Delta \mathbf{p}_{\mathrm{des}},\Delta \mathbf{\bar{q}}_{\mathrm{des}},\mathbf{p}_\mathrm{{obs}},\mathbf{dim}_{\mathrm{obs}}]$ contains the desired landing position ($\Delta \mathbf{p}_{\mathrm{des}} \in \mathbb{R}^3$), desired landing orientation ($\Delta \mathbf{\bar{q}}_{\mathrm{des}} \in \mathbb{R}^4$), the centre of the obstacle ($\mathbf{p}_{\mathrm{obs}} \in \mathbb{R}^3$
if present), and its dimensions ($\mathbf{dim}_{\mathrm{obs}} \in \mathbb{R}^3$ 
including height, width, and length)\footnote{In the training process, we sample the landing pose and obtain the obstacle parameters from the simulator. In the real world, the command vector is specified by the user.}. Due to the lack of long-term memory in the feed-forward neural network, we use the jump toggle $j$ to indicate whether the robot has already jumped, similar to 
 \cite{peng_deepmimic_2018}. However, in our case, the jump toggle also serves as a control switch, where the robot remains standing until its value is changed. 

\textbf{Action space:} 
Our policy generates the twelve actuated joint angles ($\mathbf{q}^{\mathrm{des}} \in \mathbb{R}^{12}$) for jumping control. Particularly, we learn the deviations from the nominal joint positions $\mathbf{q}^{\mathrm{nom}} \in \mathbb{R}^{12}$. 
To smooth the output actions, we used an exponential moving average (EMA) low-pass filter with a cut-off frequency of 5 Hz. The filtered actions are then scaled and added to $\mathbf{q}^{\mathrm{nom}}$ to generate $\mathbf{q}^{\mathrm{des}}$ for the motor servos, i.e. $\mathbf{q}^{\mathrm{des}} = \mathbf{a} + \mathbf{q}^{\mathrm{nom}}$. A PD feedback controller then produces the desired torque at a higher frequency, as shown in Fig. \ref{fig:controller}. To guarantee safety, we clip $\mathbf{q}^{\mathrm{des}}$ within the feasibility range when the real joint angles approach the limits.


%
\subsection{Rewards}
\bgroup
\def\arraystretch{1.4}
\begin{table*}[t]
\caption{Rewards definition. The light orange colour indicates task-based rewards, while the light purple shade describes regularisation rewards. $w_{\times}$ is the weight, $\sigma_{\times}$ is a scaling factor for the exponential kernel, $\operatorname{e}(\cdot)$ and $\operatorname{log}(\cdot)$ separately denote the exponent and logarithm operation.}
\vspace{-2mm}
\resizebox{\textwidth}{!}{
\begin{tabular}{|l|l|l|l|l|}
\hline
\rowcolor{black!5} 
 Name & Type & Stance & Flight & Landing \tabularnewline
    \hline
\rowcolor{orange!10} 
  Landing position & Single & 0 & 0 & $w_\mathbf{p}(e(-\sum||\mathbf{p}_{\mathrm{land}} - \mathbf{p}_{\mathrm{des}}||^2)/\sigma_{p\mathrm{,land}})$  \tabularnewline
    \hline
\rowcolor{orange!10} 
 Landing orientation & Single & 0 & 0 & $w_{\mathrm{ori}}(\operatorname{e}(-||\operatorname{log}(\mathbf{\bar{q}}_{\mathrm{land}}^{-1}*\mathbf{\bar{q}}_{\mathrm{des}}||^2)/\sigma_{\mathrm{ori,land}})$  \tabularnewline
    \hline
\rowcolor{orange!10} 
  Max height  & Single & 0 & 0 & $w_{h} (\operatorname{e}(||h_{\mathrm{max}} - 0.9||^2)/\sigma_{p_z\mathrm{,max}}))$  \tabularnewline
    \hline
\rowcolor{orange!10} 
  Jumping &  Single & 0 & 0 & $w_\mathrm{jump}$\tabularnewline
    \hline
\rowcolor{orange!10} 
   Base Position & Continuous &  $w_{p_z,\mathrm{st}} (\operatorname{e}(-||p_z - 0.20||^2/\sigma_{p_z,\mathrm{st)}}))$ & 
    $w_{p_z,\mathrm{fl}}(\operatorname{e}(-||p_z - 0.7||^2/\sigma_{p_z\mathrm{,fl}}))$ & $w_{\mathbf{p},\mathrm{l}} (\operatorname{e}(-\sum||\mathbf{p} - \mathbf{p}_{\mathrm{des}}||^2/\sigma_{p\mathrm{,l}}))$  \tabularnewline
    \hline
\rowcolor{orange!10} 
 Orientation Tracking & Continuous & $w_{\mathrm{ori,st}} (\operatorname{e}(-||\log(\mathbf{\bar{q}}_{\mathrm{base}}^{-1}*\mathbf{\bar{q}}_{\mathrm{des}}||^2/\sigma_{\mathrm{ori,st}}))$ & 0 & $w_{\mathrm{ori,l}} (\operatorname{e}(-||\log(\mathbf{\bar{q}}_{\mathrm{base}}^{-1}*\mathbf{\bar{q}}_{\mathrm{des}})||^2/\sigma_{\mathrm{ori,l}}))$ \tabularnewline
    \hline
\rowcolor{orange!10} 
    Base linear velocity & Continuous & 0 & $w_{\mathbf{v}_{x,y}} (-\operatorname{e}(\sum||\mathbf{v}_{x,y} - \mathbf{v}_{\mathrm{des}}||^2/\sigma_{v}))$ & 0 \tabularnewline
    \hline
\rowcolor{orange!10} 
    Base angular velocity & Continuous & 0 & $w_{\boldsymbol{\omega}} (\operatorname{e}(-\sum||\boldsymbol{\omega} - \boldsymbol{\omega}_{\mathrm{des}}||^2/\sigma_{\omega}))$ & $0.1w_{\boldsymbol{\omega}} (\operatorname{e}(-\sum||\boldsymbol{\omega}||^2/\sigma_{\omega}))$ \tabularnewline
    \hline
\rowcolor{orange!10} 
    Feet clearance & Continuous & 0 & $w_{\mathrm{feet}} (||p_{\mathrm{feet}} - p_{\mathrm{feet}}^0 + [0.0,0.0,-0.15]||^2) $ & 0 \tabularnewline
    \hline
\rowcolor{blue!10} 
    Symmetry & Continuous & \multicolumn{3}{|c|}{$w_{\mathrm{sym}}(\sum_{\mathrm{joint}}|\mathbf{q}_{\mathrm{left}} - \mathbf{q}_{\mathrm{right}}|^2)$}\tabularnewline
    \hline
\rowcolor{blue!10} 
    Nominal pose & Continuous & $w_\mathbf{q}(\operatorname{e}(-\sum_{\mathrm{joint}}||\mathbf{q}_j - \mathbf{q}_{j,\mathrm{nom}}||^2/\sigma_{q})$ & 0.1$w_\mathbf{q}(\operatorname{e}(-\sum_{\mathrm{joint}}||\mathbf{q}_j - \mathbf{q}_{j,\mathrm{nom}}||^2/\sigma_{q})$ & $w_\mathbf{q}(\operatorname{e}(-\sum_{\mathrm{joint}}||\mathbf{q}_j - \mathbf{q}_{j,\mathrm{nom}}||^2/\sigma_{q})$ \tabularnewline
    \hline
\rowcolor{blue!10} 
    Energy & Continuous & \multicolumn{3}{|c|}{$w_{\mathrm{energy}} (\boldsymbol{\tau}^T \mathbf{\dot{q}})$}\tabularnewline
    \hline
\rowcolor{blue!10} 
    Base acceleration & Continuous & \multicolumn{3}{|c|}{$w_{\mathrm{acc}} |\mathbf{\dot{v}}|^2$}\tabularnewline
    \hline
\rowcolor{blue!10} 
    Contact change & Continuous &\multicolumn{3}{|c|}{$w_{c}\sum_{\mathrm{feet}}(c_{\mathrm{foot}}(t)-c_{\mathrm{foot}}(t-1))$} \tabularnewline
    \hline
\rowcolor{blue!10} 
    Maintain Contact & Continuous & $w_{\mathrm{contact}}\sum_{\mathrm{feet}}c_{\mathrm{foot}}(t)$ & 0 & 0 \tabularnewline
    \hline
\rowcolor{blue!10}
    Contact forces & Continuous & \multicolumn{3}{|c|}{$w_{F_c}\sum_{i=0}^{n_{\mathrm{f}}}|F_i-\bar{F}|$} \tabularnewline
    \hline
\rowcolor{blue!10} 
    Action rate & Continuous & \multicolumn{3}{|c|}{$w_{a}\sum_{\mathrm{joint}}|\mathbf{a}(t) - \mathbf{a}(t-1)|^2$} \tabularnewline
    \hline
\rowcolor{blue!10} 
    Joint acceleration & Continuous & \multicolumn{3}{|c|}{$w_{\ddot{q}}\sum_{\mathrm{joint}}|\mathbf{\ddot{q}}_j|^2$} \tabularnewline
    \hline
\rowcolor{blue!10} 
    Joint limits & Continuous & \multicolumn{3}{|c|}{$w_{q_{lim}}\sum_{\mathrm{joint}}|\mathbf{q}_j-\mathbf{q}_{j,lim}|^2$} \tabularnewline
    \hline
    
\end{tabular}}
\label{tab:rewards}
\end{table*}
\egroup
Ideally, we expect the agent to accomplish the task while maximising the rewards it receives. However, poor choice of reward scaling could lead the agent to converge to the local minima, e.g., standing behaviour without jumping, where only certain penalties like energy cost and joint acceleration are minimised. 
To avoid this, 
instead of naively summing them, we multiply the positive component of the reward by the exponent of the squared negative component, i.e. $r_{\mathrm{total}} = r^{+}\operatorname{e}(-||r^{-}||^2/\sigma)$\footnote{For conciseness, notation $\operatorname{e}(-||x||^2/\sigma)$ is used to represent passing the squared error $||x||^2$ through an exponential kernel of the form $\operatorname{exp}(\frac{-||x||^2}{\sigma})$. This ensures the reward is positive and scales it between 0 and 1.}. This allows the agent to always receive a strictly positive reward, scaled down by the amount of penalties, which improves the learning stability. 

As listed in Table~\ref{tab:rewards}, three phases are used to describe when each reward is given. In particular, `stance' indicates that the robot has been given a command to jump but is still on the ground. Then, `flight' is triggered when the robot is in mid-air and has no contact with the ground. Finally, the `landing' begins upon landing and lasts until the end of the episode. 
In each phase, task-based rewards (in orange) and regularisation rewards (in violet) are considered. On the other hand, the rewards items can be divided into \textit{Single} type and \textit{Continuous} type, where the former is given once per episode (typically at the end), and the latter is given once per each simulation step that satisfies the conditions. 

\textbf{Task rewards:} First, sparse rewards are introduced to encourage the general behaviour for accomplishing the desired jumping task, including those of detecting contact (`landing') after several steps of no contact (`flight'), the maximum height the agent reached, and whether it has landed at the desired position with the desired orientation. These rewards are only given once at the end of the episode, marked by `Single' in Table~\ref{tab:rewards}. 
In addition, continuous task-related objectives are also defined to simplify the exploration, including
\begin{itemize}
    \item Tracking the desired linear velocity ($\mathbf{v}^b_{x,y, \mathrm{des}}$) and yaw angular velocity while in flight, and tracking zero angular velocity after landing.
    \item Squatting down to a height of 0.2m while on the ground and tracking a certain height in the air.
    \item Maintaining a constant base position and tracking the desired orientation after landing.
\end{itemize}

Notably, in order to ensure enough clearance when jumping forward and over obstacles, we introduce a foot clearance reward that tracks the nominal foot position (i.e. at the nominal joint angles $q^{\mathrm{nom}}$) on the xy-plane, and simultaneously, minimises the z-distance between each foot and the centre of mass. This objective encourages the robot to tuck its legs in close to its body while in the air. 

 \textbf{Regularisation rewards:} As we do not imprint any reference motions onto the agent, auxiliary regularisation rewards are needed to achieve smooth, feasible and safe behaviour. Specifically, we penalise the action rate, together with any violations of predefined soft limits for the joint position. Besides, the instantaneous energy power, computed as the dot product between actuator torque and joint velocity, is penalised for generating an energy-efficient motion. Considering that various quadrupedal jumps seen in nature exhibit high left- and right-side symmetry, we drive the robot towards maintaining this symmetry with an additional reward. Finally, we noticed that the robot often stomped its feet rapidly during the squat-down stage in the training process. 
 To eliminate this unnecessary behaviour, we add a small reward for maintaining contact in the first few steps of the episode, as well as a penalty on frequent contact state changes.

\textbf{Termination:} We terminate each episode when the following events occur:
\begin{itemize}
    \item Collision between body links and the environment.
    \item Base height lower than 0.12 m.
    \item Orientation error larger than 3.0 rad.
    \item Landing position error bigger than 0.15 m.
\end{itemize}

%
\subsection{Domain randomisation}
\label{sec:domain_rand}
\bgroup
\def\arraystretch{1.5}
\begin{table}[t]
\centering
\caption{Randomised variables and their ranges.}
\vspace{-2mm}
\begin{tabular}{|l|l|}
\hline
\rowcolor{black!20} 
 Name  & Randomisation range \tabularnewline
    \hline
Ground friction & [0.01, 3.0]  \tabularnewline
    \hline
\rowcolor{black!10} 
Ground restitution & [0.0, 0.4]  \tabularnewline
    \hline
Additional payload & [-1.0, 3.0] \si{\kilo\gram} \tabularnewline
    \hline
\rowcolor{black!10} 
Link mass factor & [0.7,1.3] x \tabularnewline
    \hline
Centre of mass displacement & [-0.1, 0.1] \si{\metre} \tabularnewline
    \hline
\rowcolor{black!10} 
Episodic Latency & [0.0, 40.0] \si{\milli\second} \tabularnewline
    \hline
Extra per-step latency & [-5.0, 5.0] \si{\milli\second}  \tabularnewline
    \hline
\rowcolor{black!10} 
Motor Strength factor & [0.9, 1.1] x \tabularnewline
    \hline
Joint offsets & [-0.02, 0.02] \si{\radian}\tabularnewline
    \hline
\rowcolor{black!10} 
PD Gains factor & [0.9, 1.1] x \tabularnewline
    \hline
Joint friction & [0.0, 0.04] \tabularnewline
    \hline
\rowcolor{black!10} 
Joint damping & [0.0, 0.01] $\si{\newton\metre\second\per\radian}$ \tabularnewline
    \hline
\end{tabular}
\label{tab:domain_randomisation}
\end{table}
\egroup
%
To bridge the gap between simulation and real-world scenarios, we implement zero-shot domain randomisation. The ground friction, restitution, and link mass are sampled at random at the start of every episode. In addition, we add a random offset to the joint encoder values, 
randomise proportional and derivative gains of the PD controller 
and randomise the strength of the motors for every episode. 
The range of each randomized variable is listed in Table~\ref{tab:domain_randomisation}. 

For hardware control, unmodelled communication delays and latencies strongly weaken the performance of learning-based policies. 
To tackle this issue, at the beginning of each episode, we sample a latency value from the range of $l \in [0,50]$ ms. Then, at each step, we add a small random value to reflect the effect of stochastic communication delays.

\section{Experimental validation}\label{experiment_validation}
In this section, we first validate the policy trained on the first two curriculum stages (i.e. policy $\pi_{II}$, shown in Fig. \ref{fig:curriculum}), through various experiments - forward and diagonal jumps, continuous jumps, and robust jumping in the presence of environmental disturbances and uneven terrains. Then, we validate the policy after the final training stage (policy $\pi_{III}$) when jumping onto and over obstacles. 
\\
\subsection{Training setup}
The
implementation is based on the open-source Gym environment provided by ETH Zurich \cite{rudin_learning_2022}. Particularly, we use 4096 agents and 24 environmental steps per agent per update step. 
For the vertical jump, we train for 3k iterations, while for the forward jump without and with obstacles we train for 10k steps each. 
 The actor policy and the critic are parameterised by a shared MLP with 3 hidden layers of dimensions $[256, 128, 64]$, with exponential linear unit (ELU) activations after each layer. Using a single RTX 3090 GPU, the three highly parallelised training stages took approximately 1.4 hours, 4.1 hours and 4.8 hours, respectively.

 The policy operates at a frequency of 50 Hz, and the simulation runs at 200 Hz.
We performed all of the experiments on the Unitree Go1. During the hardware validations, we use a constant joint friction value of 0.04, joint damping of 0.01 Nm s rad$^{-1}$ and a constant latency of 30 ms.

\subsection{Versatile jumping on flat ground}
 
\subsubsection{Forward jumping}
\begin{figure*}
    \centering
    \includegraphics[width=\textwidth]{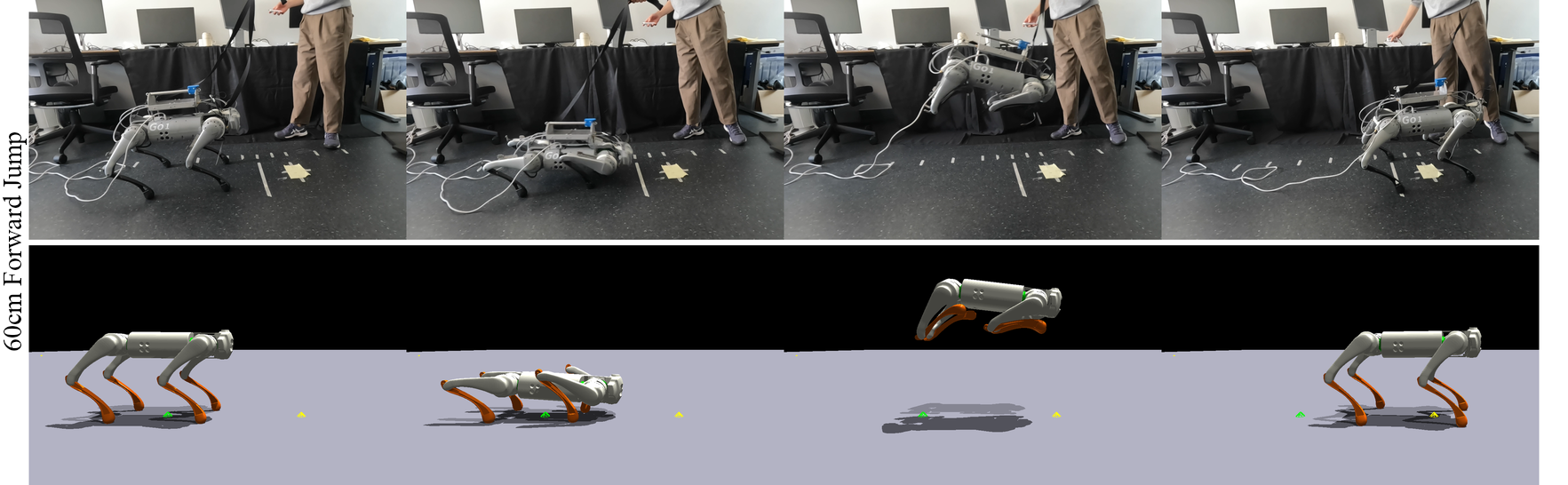}
    \caption{Real world (top) and simulation (bottom) execution of a forward jump. The yellow marker indicates the desired 60cm jumping distance.}
    \label{fig:forward_jump_60cm}
\end{figure*}
\begin{figure*}
\subfloat[Joint angles, velocities and torques for the front right (FR) leg during the 60cm forward jump. 
The flight phase for the hardware experiment is indicated by the yellow-shaded region.\label{fig:forward_jump_60cm_data_1}]{%
  \includegraphics[clip,width=\textwidth]{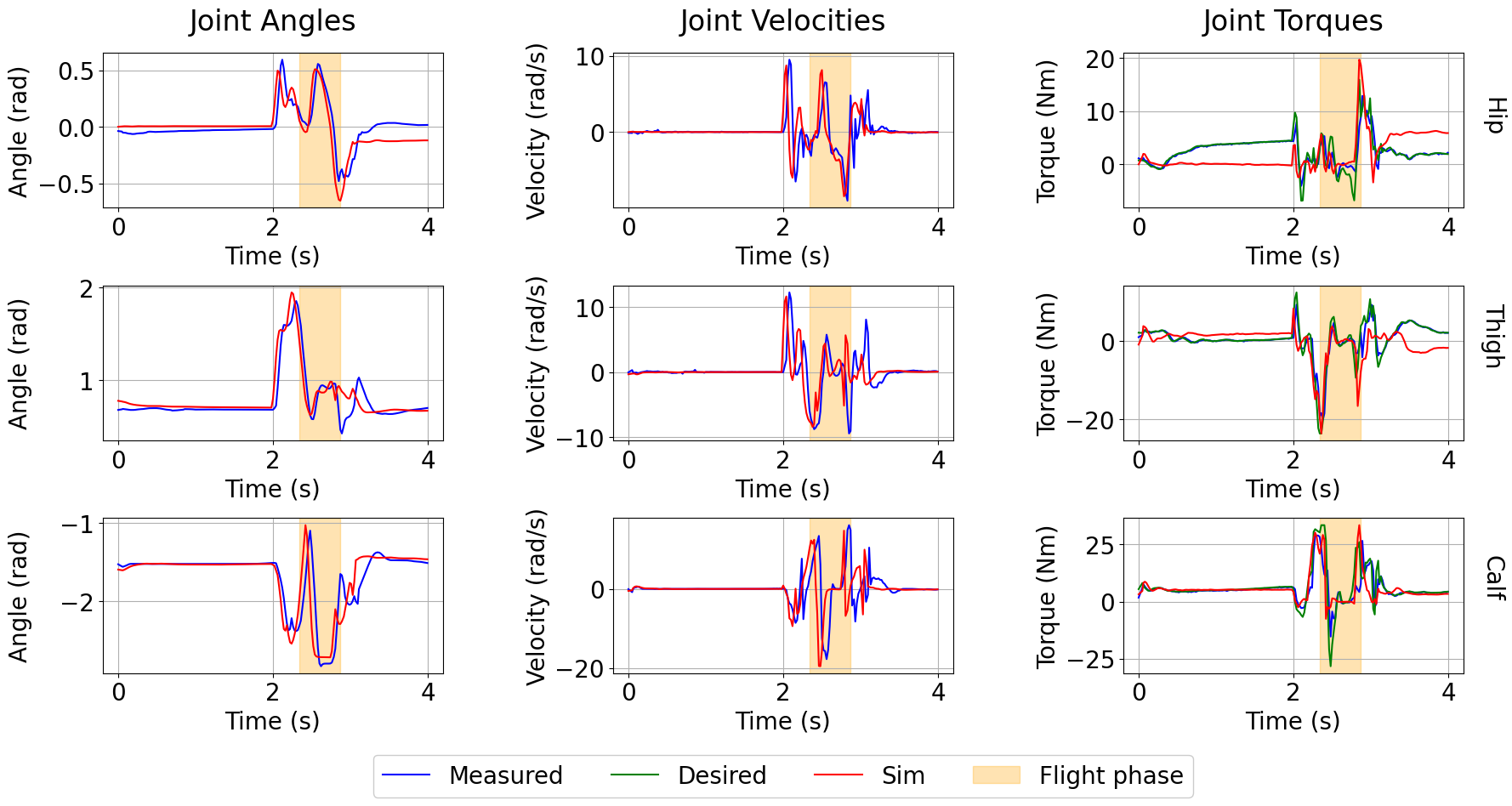}%
}
\\
\subfloat[Base angular and linear velocity during the 60cm forward jump. 
The flight phase for the hardware test is indicated in light yellow.\label{fig:forward_jump_60cm_data_2}]{\includegraphics[clip,width=\textwidth]{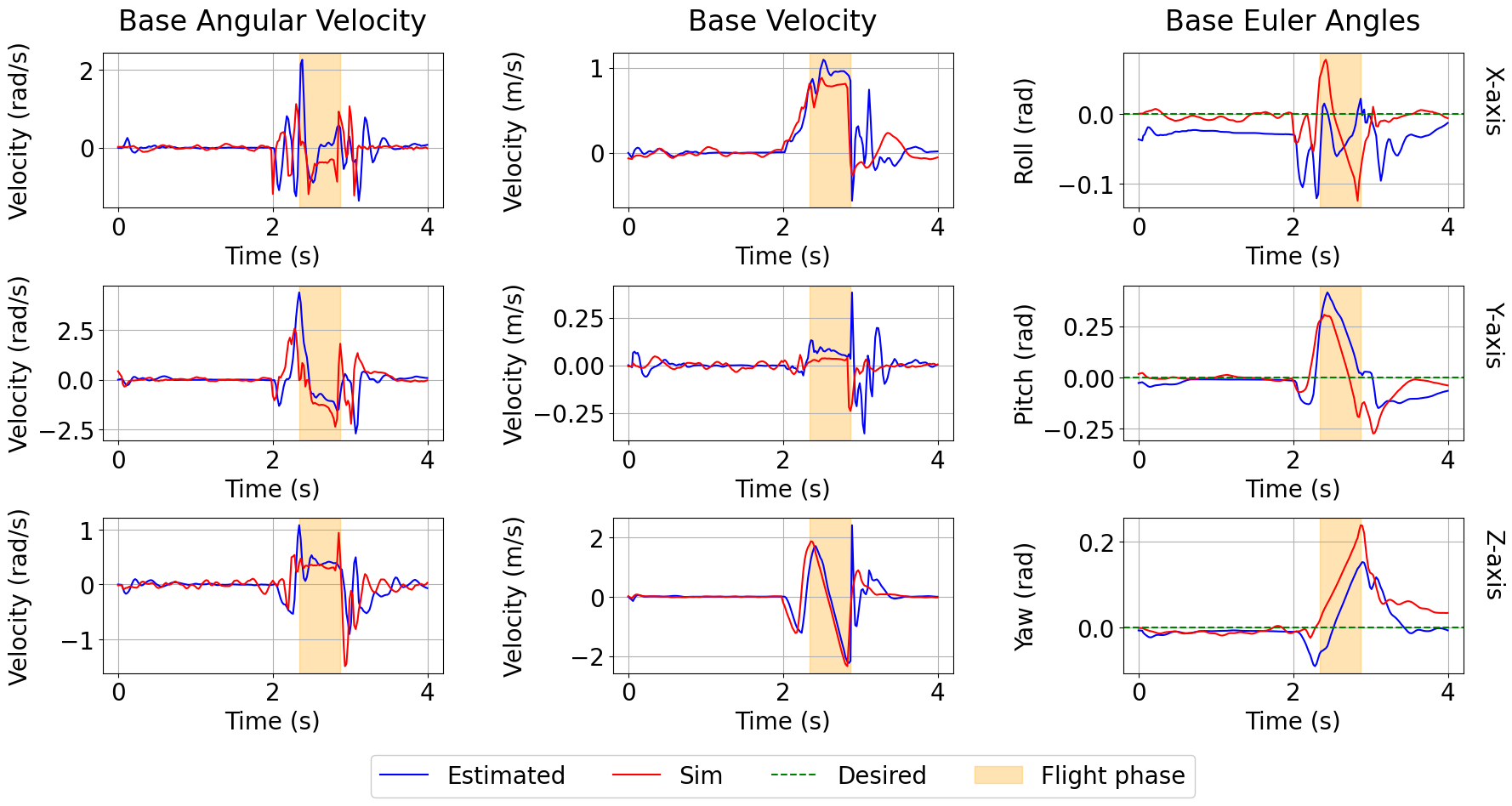}%
}
\caption{Hardware and simulation quantitative results for the 60cm forward jump.}
\label{fig:hardware_vs_sim_60cm}
\end{figure*}
First, we evaluate the policy on a variety of forward jumps. Fig.~\ref{fig:forward_jump_60cm} compares hardware and simulation motions of a 60cm forward jump while Fig.~\ref{fig:hardware_vs_sim_60cm} presents the quantitative results.  As can be seen, the real-world behaviour closely matches the simulated prediction. One noticeable deviation is in the peak torques at take-off - where the measured torques deviate from both the desired torques (computed by the PD control law using the desired joint angles) and the simulation torques. 
Besides, larger joint angles for the hip and thigh are measured upon landing in real-world tests, likely due to poor impact modelling in the simulation. Finally, the Euler angles show a slight variation between simulation and hardware. We believe that this mismatch is mainly due to the motor modelling inaccuracies, coupled with the weight of the additional mass on top of the robot, shifting its centre of mass. Despite these state deviations, the jumping distance is well-tracked, and the base velocity matches the expected behaviour, 
showing a good sim2real adaptation. 
%
%
\begin{figure*}
    \includegraphics[width=\textwidth]{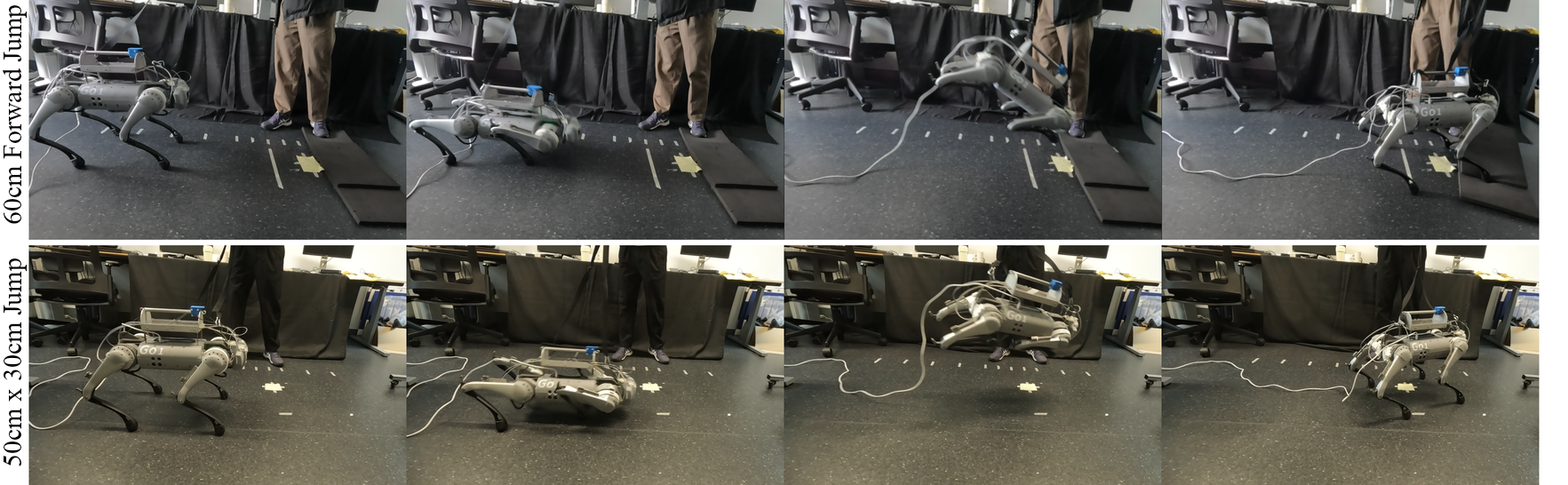}
    \vspace{-6mm}
    \caption{Hardware results for a 90cm forward jump (top) and a 50cm $\times$ 30cm diagonal jump with desired yaw of $30 \degree$ (bottom).}
    \label{fig:forward_diagonal}
\end{figure*}

We then tested the maximum distance it could jump across. Fig.~\ref{fig:forward_diagonal}(top) illustrates a 90 cm forward jump, with the target landing point shown by the yellow marker. 
Despite slipping on the soft pads as it lands, the robot recovers quickly, demonstrating its robustness against uncertainties\footnote{It is worth mentioning that we reward the position of the base upon landing, rather than the feet. As a result, in the trial, the base cleared the 90cm distance, but the rear left foot landed a bit behind.}. To the best of our knowledge, this is the largest jumping distance achieved by robots of similar size and similar actuators (see Table \ref{tab:comparison_methods}). 

\begin{table}[]
    \caption{Maximal jump length comparison with state-of-the-art. 
    }
    \centering
    \begin{tabular}{c|c|c|c|c|c|c|c}
    \hline
        {Method} & 
         \cite{smith_learning_2023}  & \cite{margolis_learning_2021} &  \cite{caluwaerts_barkour_2023} & \cite{yang_continuous_2023} &
         \cite{yang_cajun_2023}&\cite{cheng_extreme_2023} & \textbf{Ours}  \\
        \hline
        {Jump length [m]} & 0.2 & 0.26 & 0.5& 0.6 & 0.7 & 0.8 & \textbf{0.9}\\
        \hline
    \end{tabular}
    \label{tab:comparison_methods}
\end{table}

\subsubsection{Diagonal jumping}
%
 Fig.~\ref{fig:forward_diagonal}(bottom) shows a diagonal jump of 50 cm x 30 cm with a desired yaw of $30\degree$. Both the landing position and yaw are tracked accurately. 
%
\begin{figure*}
    \centering
    \includegraphics[width=\textwidth]{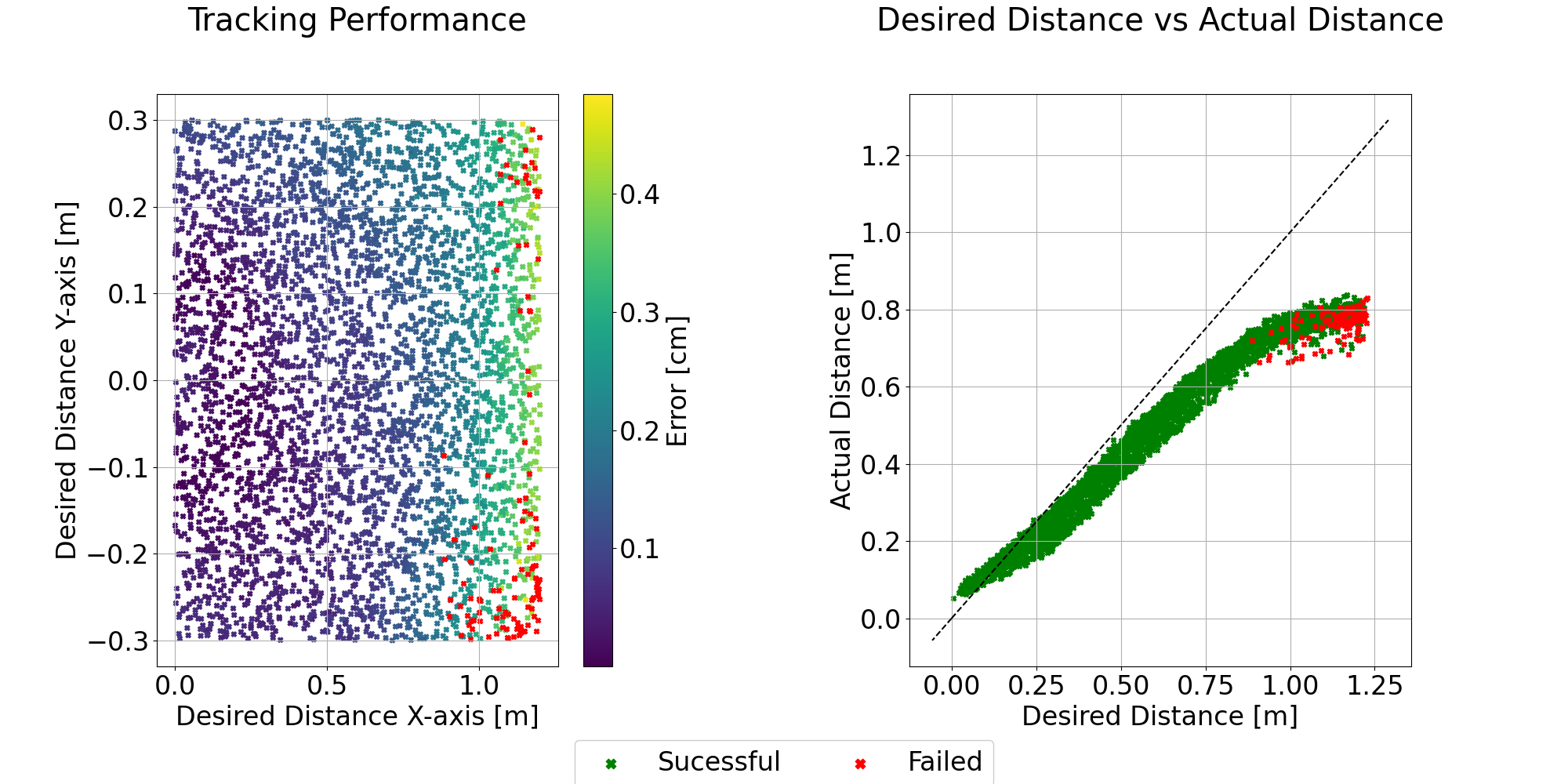}
    \caption{Tracking performance as a function of the desired X- and Y-axis jumping distances, with the error (in cm) shown by the colour gradient (left); and the tracking performance in terms of overall desired vs actual jumping distance (right). The environments that have been terminated (due to any non-foot collisions) are shown in red, and the black $45\degree$ dashed line indicates the ideal tracking performance. Data is gathered from 8000 trials across the whole jumping range $x\in [0,1.2],y \in [-0.3,0.3]$. Only 112 robots have been terminated, leading to a success rate of 98.6\%.}
    \label{fig:forward_success}
\end{figure*}
\\Furthermore, we evaluated the policy across the whole jumping range in simulation, of which the success rate and tracking metrics are presented in Fig.~\ref{fig:forward_success}. As can be seen from the left plot, the tracking error is lowest for narrow jumps of forward distance up to 50 cm. As both the longitudinal and lateral distances increase, so does the final landing error. Interestingly, the majority of failed environments asymmetrically occur in the lower right corner of the plot. The right plot in Fig. \ref{fig:forward_success} shows the same data but grouped by total desired distance vs actual achieved distance. We found that the data closely follow the $45\degree$ line (i.e. ideal performance) for the smaller jumps with the gradient slowly decreasing after 50 cm.
\subsection{Jumping onto/across rough terrain}
\begin{figure*}
    \centering
    \includegraphics[width=\textwidth]{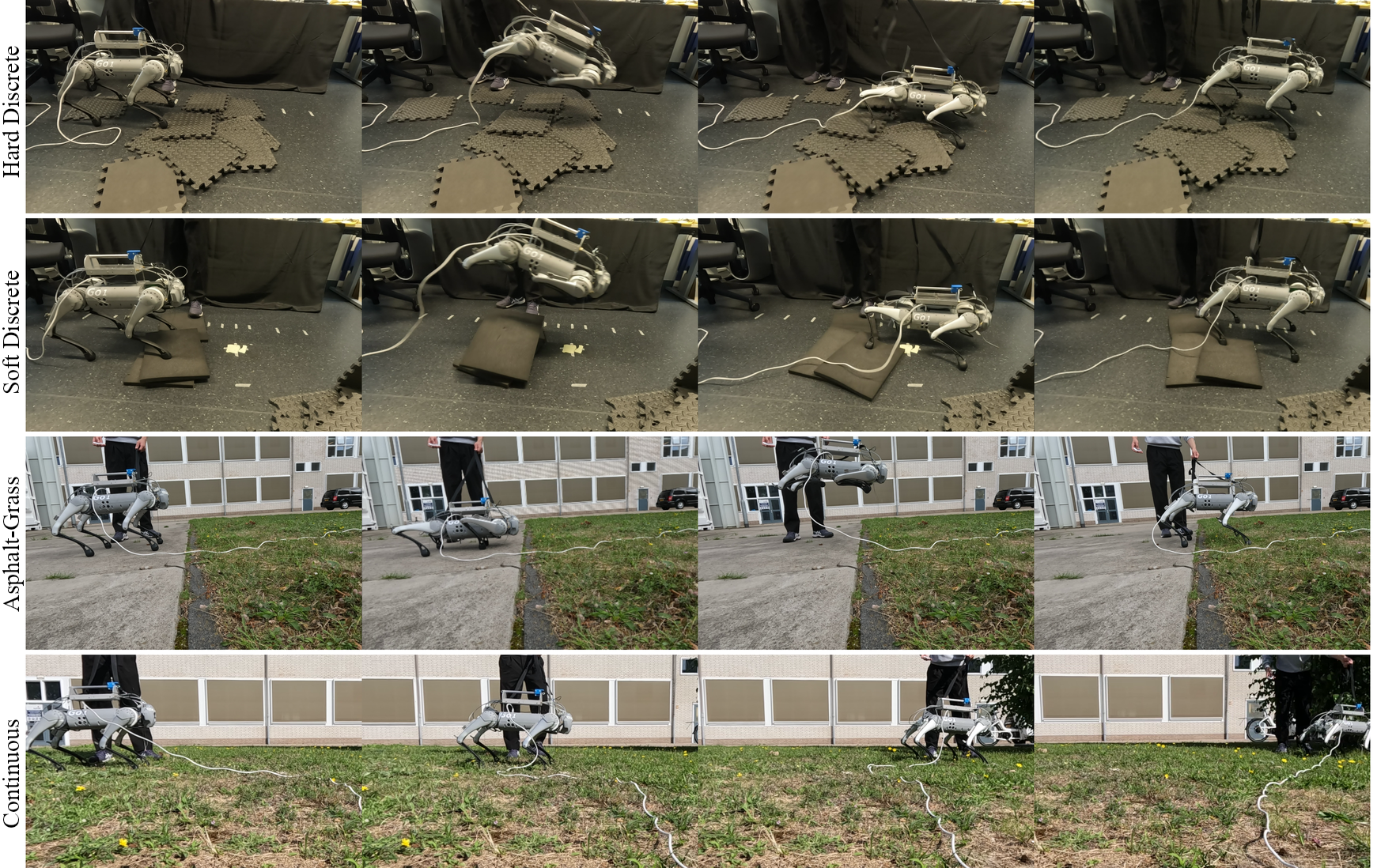}
    \vspace{-8mm}
    \caption{Several experiments showcasing the robustness of the policy $\pi_{II}$ to variations in the terrain: jumping across discrete hard and soft objects (rows 1 and 2), asphalt-to-grass jump (row 3), nine consecutive jump on grass (row 4). }
    \label{fig:rough_terrain}
\end{figure*}

We here evaluate how well the policy performs in the presence of environmental disturbances, despite not being trained on uneven or rough ground. In this section, we ran several experiments, including jumping with obstacles surrounding the robot, blindly jumping from and onto a box, and jumping from asphalt onto a soft grassy terrain.
As shown by the top two time-lapses in Fig.~\ref{fig:rough_terrain}, the policy enables robust jumping onto both soft and stiff objects that could (and did) slip under the feet of the robot. The third row demonstrates that the robot could jump from hard asphalt onto soft grass, despite training on flat ground only. 

Next, we tested the policy on a continuous jumping task, where a new command of a 40cm forward jump is given following each jump without resetting the robot states. 
As seen in the fourth row of Fig.~\ref{fig:rough_terrain}, the policy is robust enough to execute a jump from a variety of different initial states. Despite the fact that the soft ground causes some hip angle deviation upon landing, the robot was able to execute at least nine consecutive jumps.
%
\subsection{Forward jumping with obstacles}
\begin{figure*}
    \centering
    \includegraphics[width=\textwidth]{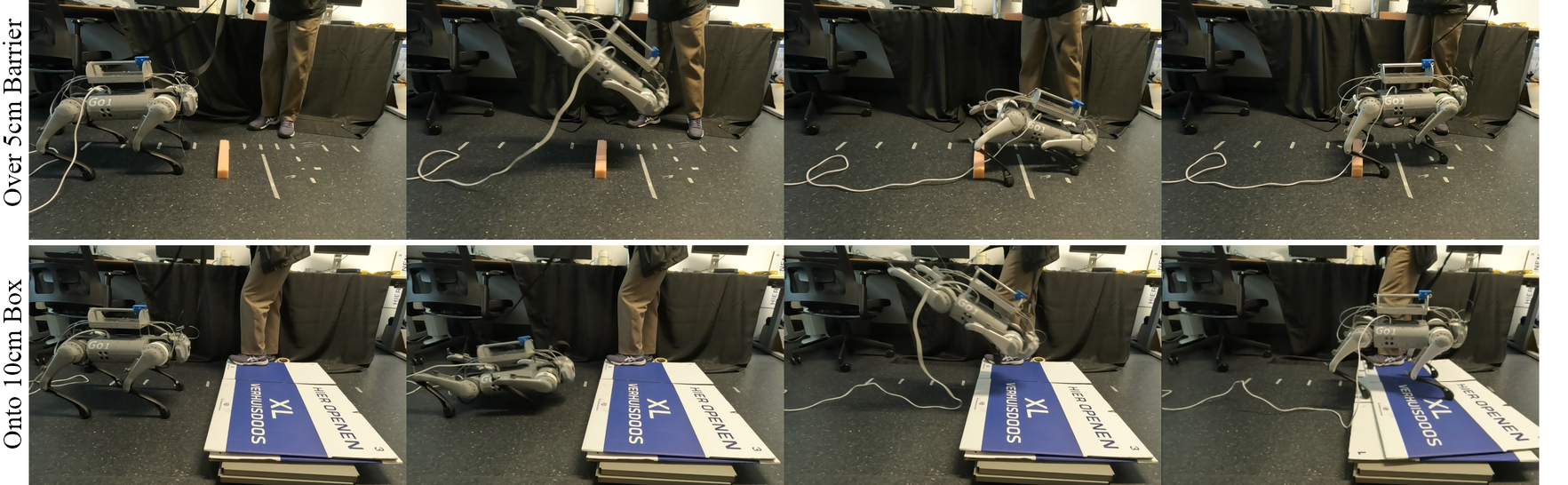}
    \vspace{-8mm}
    \caption{Jumping over a 5cm tall, 5cm wide obstacle (top row) and jumping onto a 10cm tall box (bottom). }
    \label{fig:obstacles_hardware}
\end{figure*}
To further demonstrate the versatility, we tested forward jumping with obstacles, using policy $\pi_{III}$. To be brief, only two scenarios are presented here, including jumping over a 5 cm tall thin obstacle and landing on a 10 cm box. 
In the first task, the robot had to jump across 80 cm to avoid collision. As seen in the top row of Fig.~\ref{fig:obstacles_hardware}, the robot succeeded in jumping over the barrier and landed successfully. 
In the second case, the robot needed to leap over 70 cm while maintaining a large height. As a result, better performance was observed, considering that a shorter forward distance enabled the robot to achieve a larger height throughout the flight. 
\subsection{Ablation study} \label{sec:ablation_study}
\begin{figure}
\subfloat[\label{fig:upwards_ablation}]{%
  \includegraphics[clip,width=\columnwidth]{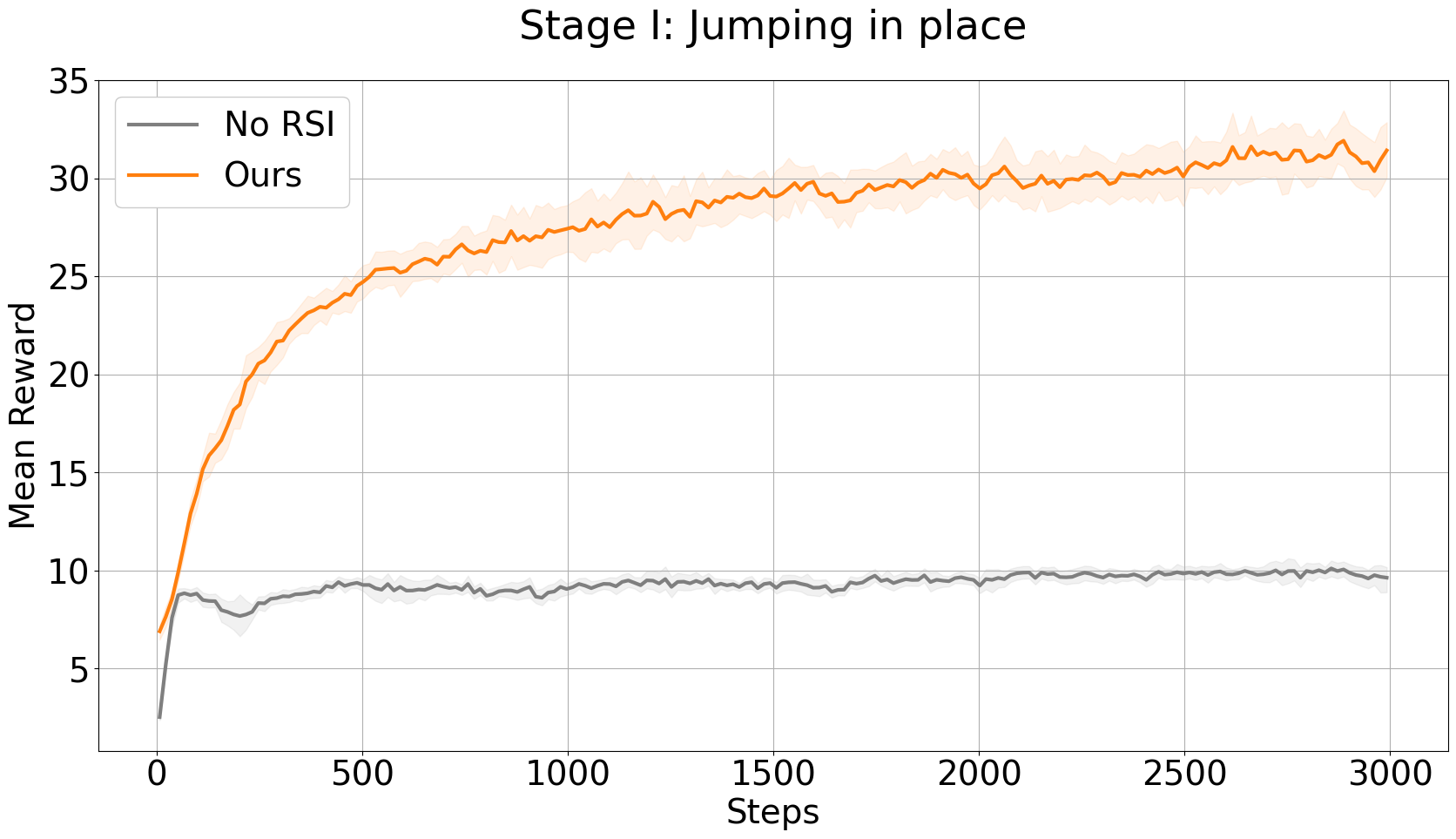}%
}
\\
\subfloat[\label{fig:forward_ablation}]{\includegraphics[clip,width=\columnwidth]{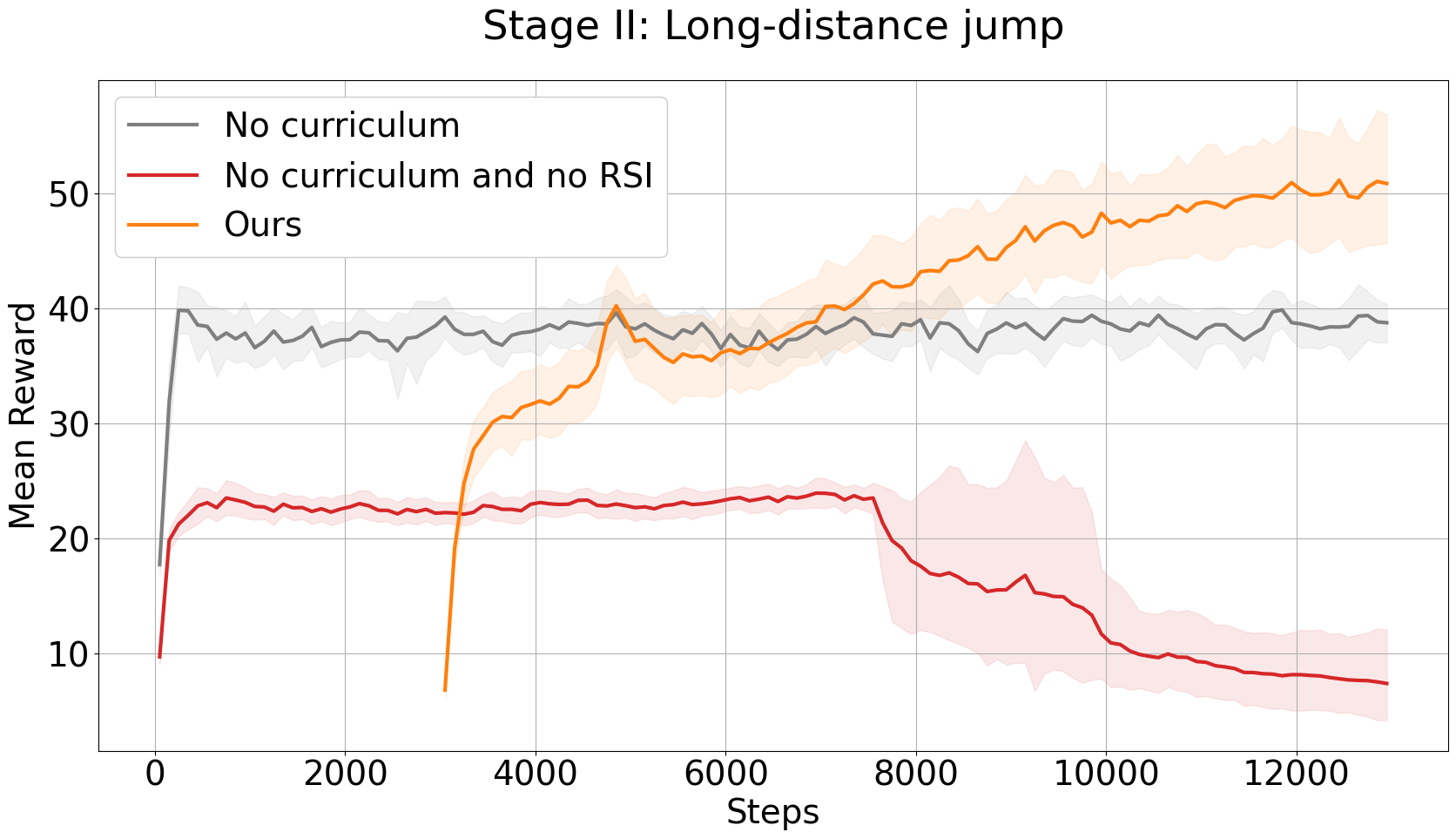}%
}
\caption{Mean reward throughout training for the (a) Stage I: Jumping in place, and (b) Stage II: Long-distance jump tasks. 
}
\label{fig:ablation_study}
\end{figure}

To better understand the effect of our curriculum, we compared our approach to several baselines in Fig.~\ref{fig:ablation_study}:
\begin{itemize}
    \item \textbf{No RSI}: Training Stage I 
    without RSI, i.e. no height and upward velocity initialisation,
    \item \textbf{No curriculum}: Directly training Stage II 
    without pre-training Stage I, but with RSI height and velocity initialisation. For fairness, we train this baseline for an additional 3k steps, 
    \item \textbf{No curriculum and no RSI}: Same as above, but without any RSI.
\end{itemize}

As can be seen from Fig.~\ref{fig:upwards_ablation}, the RSI is required for learning the jumping-in-place task. Without it, the agent converges to a local optimum 
and fails to complete the task. Despite the overall high reward, it can be seen in Fig. \ref{fig:forward_ablation} that directly training the long-distance jump also results in an early convergence to a standing behaviour, which highlights the need for our curriculum strategy. 

\section{Discussion and conclusion}\label{conclusion_part}
In this work, we present a curriculum-based end-to-end deep reinforcement learning approach, capable of learning a variety of precise short- and long-distance jumps, while also reaching the desired yaw upon landing. Unlike many existing methods, we have achieved this through a single policy, without the need for reference trajectories and additional imitation rewards. 
Furthermore, through domain randomisation, we successfully deployed the policy onto the real system and closely matched the expected behaviour from the simulation. The system was robust to the noisy sensor data, especially the foot contact sensors and the velocity state estimates. The jumps exhibited high accuracy, both in simulation and on the hardware, in terms of tracking the desired landing position and orientation. Furthermore, our policy achieved a 90 cm forward jump on the Unitree Go1 robot, a distance greater than those reported by other model- and learning-based controllers. We demonstrated additional outdoor tests, where the robot successfully performed nine consecutive jumps on soft grass, without previously encountering such environments in its training. In addition, we showed that simulating obstacles throughout training and conditioning the policy on their properties can enhance the mobility of the robot, allowing it to safely leap over or land on objects of up to 10cm.

When executing a long-distance jump, real animals exhibit a four-legged contact phase, followed by an upward pitch and pure rear-leg contact at take-off. During landing a mirrored behaviour is observed - the body is pitching downwards and contact is first gained with the front legs. Previous model-based control works \cite{nguyen_optimized_2019,nguyen_contact-timing_2022} have manually incorporated this contact schedule into their optimisers. It would be interesting to investigate how such behaviour can be learned through DRL without supplying a reference trajectory, and validate its benefits compared to the style of jumping exhibited here. 


%
%

\printbibliography 

@misc{kumar_rma_2021,
	title = {{RMA}: {Rapid} {Motor} {Adaptation} for {Legged} {Robots}},
	shorttitle = {{RMA}},
	url = {http://arxiv.org/abs/2107.04034},
	doi = {10.48550/arXiv.2107.04034},
	urldate = {2022-10-17},
	publisher = {arXiv},
	author = {Kumar, Ashish and Fu, Zipeng and Pathak, Deepak and Malik, Jitendra},
	month = jul,
	year = {2021},
	note = {arXiv:2107.04034 [cs]},
}

@misc{agarwal_legged_2022,
	title = {Legged {Locomotion} in {Challenging} {Terrains} using {Egocentric} {Vision}},
	url = {http://arxiv.org/abs/2211.07638},
	doi = {10.48550/arXiv.2211.07638},
	urldate = {2022-11-16},
	publisher = {arXiv},
	author = {Agarwal, Ananye and Kumar, Ashish and Malik, Jitendra and Pathak, Deepak},
	month = nov,
	year = {2022},
	note = {arXiv:2211.07638 [cs, eess]},
}

@article{bjelonic2022offline,
  title={Offline motion libraries and online MPC for advanced mobility skills},
  author={Bjelonic, Marko and Grandia, Ruben and Geilinger, Moritz and Harley, Oliver and Medeiros, Vivian S and Pajovic, Vuk and Jelavic, Edo and Coros, Stelian and Hutter, Marco},
  journal={The International Journal of Robotics Research},
  volume={41},
  number={9-10},
  pages={903--924},
  year={2022},
  publisher={SAGE Publications Sage UK: London, England}
}

@article{ding2023robust,
  title={Robust Jumping with an Articulated Soft Quadruped via Trajectory Optimization and Iterative Learning},
  author={Ding, Jiatao and van L{\"o}ben Sels, Mees A and Angelini, Franco and Kober, Jens and Della Santina, Cosimo},
  journal={IEEE Robotics and Automation Letters},
  volume={9},
  number={1},
  pages={255--262},
  year={2023},
  publisher={IEEE}
}

@article{schulman2017proximal,
  title={Proximal policy optimization algorithms},
  author={Schulman, John and Wolski, Filip and Dhariwal, Prafulla and Radford, Alec and Klimov, Oleg},
  journal={arXiv preprint arXiv:1707.06347},
  year={2017}
}

@article{peng_amp_2022,
  title={Amp: Adversarial motion priors for stylized physics-based character control},
  author={Peng, Xue Bin and Ma, Ze and Abbeel, Pieter and Levine, Sergey and Kanazawa, Angjoo},
  journal={ACM Transactions on Graphics (ToG)},
  volume={40},
  number={4},
  pages={1--20},
  year={2021},
  publisher={ACM New York, NY, USA}
}

@article{peng_deepmimic_2018,
	title = {{DeepMimic}: example-guided deep reinforcement learning of physics-based character skills},
	volume = {37},
	shorttitle = {{DeepMimic}},
	pages = {143:1--143:14},
	number = {4},
	journaltitle = {{ACM} Transactions on Graphics},
	shortjournal = {{ACM} Trans. Graph.},
	author = {Peng, Xue Bin and Abbeel, Pieter and Levine, Sergey and van de Panne, Michiel},
	urldate = {2022-10-19},
	date = {2018-07},
}

@article{miki_learning_2022,
	title = {Learning robust perceptive locomotion for quadrupedal robots in the wild},
	volume = {7},
	pages = {eabk2822},
	number = {62},
	journaltitle = {Science Robotics},
	author = {Miki, Takahiro and Lee, Joonho and Hwangbo, Jemin and Wellhausen, Lorenz and Koltun, Vladlen and Hutter, Marco},
	urldate = {2022-10-13},
	date = {2022-01},
}

@article{lee_learning_2020,
	title = {Learning quadrupedal locomotion over challenging terrain},
	volume = {5},
	pages = {eabc5986},
	number = {47},
	journaltitle = {Science Robotics},
	author = {Lee, Joonho and Hwangbo, Jemin and Wellhausen, Lorenz and Koltun, Vladlen and Hutter, Marco},
	urldate = {2022-10-13},
	date = {2020-10},
}

@article{hwangbo_learning_2019,
	title = {Learning agile and dynamic motor skills for legged robots},
	volume = {4},
	pages = {eaau5872},
	number = {26},
	journaltitle = {Science Robotics},
	author = {Hwangbo, Jemin and Lee, Joonho and Dosovitskiy, Alexey and Bellicoso, Dario and Tsounis, Vassilios and Koltun, Vladlen and Hutter, Marco},
	urldate = {2022-10-13},
	date = {2019-01},
}

@misc{vezzi_two-stage_2023,
	title = {Two-Stage Learning of Highly Dynamic Motions with Rigid and Articulated Soft Quadrupeds},
	number = {{arXiv}:2309.09682},
	publisher = {{arXiv}},
	author = {Vezzi, Francecso and Ding, Jiatao and Raffin, Antonin and Kober, Jens and Della Santina, Cosimo},
	urldate = {2023-11-03},
	date = {2023-09},
	eprinttype = {arxiv},
	eprint = {2309.09682},
}

@misc{zhuang_robot_2023,
	title = {Robot Parkour Learning},
	number = {{arXiv}:2309.05665},
	publisher = {{arXiv}},
	author = {Zhuang, Ziwen and Fu, Zipeng and Wang, Jianren and Atkeson, Christopher and Schwertfeger, Soeren and Finn, Chelsea and Zhao, Hang},
	urldate = {2023-10-23},
	date = {2023-09},
	eprinttype = {arxiv},
	eprint = {2309.05665},
}

@misc{cheng_extreme_2023,
	title = {Extreme Parkour with Legged Robots},
	number = {{arXiv}:2309.14341},
	publisher = {{arXiv}},
	author = {Cheng, Xuxin and Shi, Kexin and Agarwal, Ananye and Pathak, Deepak},
	urldate = {2023-10-11},
	date = {2023-09},
	eprinttype = {arxiv},
	eprint = {2309.14341},
}

@misc{hoeller_anymal_2023,
	title = {{ANYmal} Parkour: Learning Agile Navigation for Quadrupedal Robots},
	shorttitle = {{ANYmal} Parkour},
	number = {{arXiv}:2306.14874},
	publisher = {{arXiv}},
	author = {Hoeller, David and Rudin, Nikita and Sako, Dhionis and Hutter, Marco},
	urldate = {2023-10-12},
	date = {2023-06},
	eprinttype = {arxiv},
	eprint = {2306.14874},
}

@online{caluwaerts_barkour_2023,
	title = {Barkour: Benchmarking Animal-level Agility with Quadruped Robots},
	titleaddon = {{arXiv}.org},
	author = {Caluwaerts, Ken and Iscen, Atil and Kew, J. Chase and Yu, Wenhao and Zhang, Tingnan and Freeman, Daniel and Lee, Kuang-Huei and Lee, Lisa and Saliceti, Stefano and Zhuang, Vincent and Batchelor, Nathan and Bohez, Steven and Casarini, Federico and Chen, Jose Enrique and Cortes, Omar and Coumans, Erwin and Dostmohamed, Adil and Dulac-Arnold, Gabriel and Escontrela, Alejandro and Frey, Erik and Hafner, Roland and Jain, Deepali and Jyenis, Bauyrjan and Kuang, Yuheng and Lee, Edward and Luu, Linda and Nachum, Ofir and Oslund, Ken and Powell, Jason and Reyes, Diego and Romano, Francesco and Sadeghi, Feresteh and Sloat, Ron and Tabanpour, Baruch and Zheng, Daniel and Neunert, Michael and Hadsell, Raia and Heess, Nicolas and Nori, Francesco and Seto, Jeff and Parada, Carolina and Sindhwani, Vikas and Vanhoucke, Vincent and Tan, Jie},
	urldate = {2023-06-21},
	date = {2023-05},
	langid = {english},
}

@misc{yang_cajun_2023,
	title = {{CAJun}: Continuous Adaptive Jumping using a Learned Centroidal Controller},
	shorttitle = {{CAJun}},
	number = {{arXiv}:2306.09557},
	publisher = {{arXiv}},
	author = {Yang, Yuxiang and Shi, Guanya and Meng, Xiangyun and Yu, Wenhao and Zhang, Tingnan and Tan, Jie and Boots, Byron},
	urldate = {2023-06-20},
	date = {2023-06},
	eprinttype = {arxiv},
	eprint = {2306.09557},
}

@inproceedings{yang_continuous_2023,
	title = {Continuous Versatile Jumping Using Learned Action Residuals},
	eventtitle = {Learning for Dynamics and Control Conference},
	pages = {770--782},
	booktitle = {Annual Learning for Dynamics and Control Conference},
	author = {Yang, Yuxiang and Meng, Xiangyun and Yu, Wenhao and Zhang, Tingnan and Tan, Jie and Boots, Byron},
	urldate = {2023-06-19},
	date = {2023-06},
	langid = {english},
}

@inproceedings{nguyen_contact-timing_2022
,
  title={Contact-timing and trajectory optimization for 3d jumping on quadruped robots},
  author={Nguyen, Chuong and Nguyen, Quan},
  booktitle={IEEE/RSJ International Conference on Intelligent Robots and Systems},
  pages={11994--11999},
  year={2022}
}

@inproceedings{rudin_learning_2022,
	title = {Learning to Walk in Minutes Using Massively Parallel Deep Reinforcement Learning},
	eventtitle = {Conference on Robot Learning},
	pages = {91--100},
	booktitle = {Conference on Robot Learning},
	author = {Rudin, Nikita and Hoeller, David and Reist, Philipp and Hutter, Marco},
	urldate = {2022-10-14},
	date = {2022-01},
	langid = {english},
}

@article{rudin_cat-like_2022,
	title = {Cat-like Jumping and Landing of Legged Robots in Low-gravity Using Deep Reinforcement Learning},
	volume = {38},
	pages = {317--328},
	number = {1},
	journaltitle = {{IEEE} Transactions on Robotics},
	shortjournal = {{IEEE} Trans. Robot.},
	author = {Rudin, Nikita and Kolvenbach, Hendrik and Tsounis, Vassilios and Hutter, Marco},
	urldate = {2022-09-30},
	date = {2022-02},
}

@misc{vollenweider_advanced_2022,
	title = {Advanced Skills through Multiple Adversarial Motion Priors in Reinforcement Learning},
	number = {{arXiv}:2203.14912},
	publisher = {{arXiv}},
	author = {Vollenweider, Eric and Bjelonic, Marko and Klemm, Victor and Rudin, Nikita and Lee, Joonho and Hutter, Marco},
	urldate = {2022-10-14},
	date = {2022-03},
	eprinttype = {arxiv},
	eprint = {2203.14912},
}

@inproceedings{nguyen_optimized_2019,
	title = {Optimized Jumping on the {MIT} Cheetah 3 Robot},
	pages = {7448--7454},
	booktitle = {International Conference on Robotics and Automation},
	author = {Nguyen, Quan and Powell, Matthew J. and Katz, Benjamin and Carlo, Jared Di and Kim, Sangbae},
	date = {2019-05},
}

@inproceedings{smith_learning_2023,
  author       = {Laura M. Smith and
                  J. Chase Kew and
                  Tianyu Li and
                  Linda Luu and
                  Xue Bin Peng and
                  Sehoon Ha and
                  Jie Tan and
                  Sergey Levine},
  title        = {Learning and Adapting Agile Locomotion Skills by Transferring Experience},
  booktitle    = {Robotics: Science and Systems XIX},
  year         = {2023}
}

@article{yu_dynamic_2022,
  author       = {Fangzhou Yu and
                  Ryan Batke and
                  Jeremy Dao and
                  Jonathan W. Hurst and
                  Kevin Green and
                  Alan Fern},
  title        = {Dynamic Bipedal Maneuvers through Sim-to-Real Reinforcement Learning},
  journal      = {CoRR},
  volume       = {abs/2207.07835},
  year         = {2022},
}

@article{li_robust_2023,
  author       = {Zhongyu Li and
                  Xue Bin Peng and
                  Pieter Abbeel and
                  Sergey Levine and
                  Glen Berseth and
                  Koushil Sreenath},
  title        = {Robust and Versatile Bipedal Jumping Control through Multi-Task Reinforcement
                  Learning},
  journal      = {CoRR},
  volume       = {abs/2302.09450},
  year         = {2023},
}

@article{bellegarda_robust_2021,
  author       = {Guillaume Bellegarda and Chuong Nguyen and
                  Quan Nguyen},
  title        = {Robust Quadruped Jumping via Deep Reinforcement Learning},
  journal      = {CoRR},
  volume       = {abs/2011.07089},
  year         = {2023},
}

@article{yin_discovering_2021,
  author       = {Zhiqi Yin and
                  Zeshi Yang and
                  Michiel van de Panne and
                  KangKang Yin},
  title        = {Discovering diverse athletic jumping strategies},
  journal      = {{ACM} Trans. Graph.},
  volume       = {40},
  number       = {4},
  pages        = {91:1--91:17},
  year         = {2021},
  url          = {https://doi.org/10.1145/3450626.3459817},
  doi          = {10.1145/3450626.3459817},
  bibsource    = {dblp computer science bibliography, https://dblp.org}
}

@inproceedings{fuchioka_opt-mimic_2022,
  author       = {Yuni Fuchioka and
                  Zhaoming Xie and
                  Michiel van de Panne},
  title        = {OPT-Mimic: Imitation of Optimized Trajectories for Dynamic Quadruped
                  Behaviors},
  booktitle    = {International Conference on Robotics and Automation},
  pages        = {5092--5098},
  year         = {2023},
}

@inproceedings{margolis_learning_2021,
  author       = {Gabriel B. Margolis and
                  Tao Chen and
                  Kartik Paigwar and
                  Xiang Fu and
                  Donghyun Kim and
                  Sangbae Kim and
                  Pulkit Agrawal},
  title        = {Learning to Jump from Pixels},
  booktitle    = {Conference on Robot Learning},
  volume       = {164},
  pages        = {1025--1034},
  publisher    = {{PMLR}},
  year         = {2021}
}

@article{xie_allsteps_2020,
	title = {{ALLSTEPS}: Curriculum-driven Learning of Stepping Stone Skills},
	volume = {39},
	shorttitle = {{ALLSTEPS}},
	pages = {213--224},
	number = {8},
	journaltitle = {Computer Graphics Forum},
	author = {Xie, Zhaoming and Ling, Hung Yu and Kim, Nam Hee and van de Panne, Michiel},
	urldate = {2022-10-18},
	date = {2020},
	langid = {english},
}

@inproceedings{li_learning_2022,
  author       = {Chenhao Li and
                  Marin Vlastelica and
                  Sebastian Blaes and
                  Jonas Frey and
                  Felix Grimminger and
                  Georg Martius},
  title        = {Learning Agile Skills via Adversarial Imitation of Rough Partial Demonstrations},
  booktitle    = {Conference on Robot Learning},
  volume       = {205},
  pages        = {342--352},
  year         = {2022},
}

@misc{escontrela_adversarial_2022,
	title = {Adversarial Motion Priors Make Good Substitutes for Complex Reward Functions},
	number = {{arXiv}:2203.15103},
	publisher = {{arXiv}},
	author = {Escontrela, Alejandro and Peng, Xue Bin and Yu, Wenhao and Zhang, Tingnan and Iscen, Atil and Goldberg, Ken and Abbeel, Pieter},
	urldate = {2022-12-01},
	date = {2022-03},
	eprinttype = {arxiv},
	eprint = {2203.15103},
}

\end{document}